 \documentclass[pmlr,twocolumn]{jmlr} 

\usepackage{pifont}
\usepackage{wrapfig}

\usepackage{booktabs}
\usepackage[utf8]{inputenc} 
\usepackage[T1]{fontenc}    
\usepackage{hyperref}       
\usepackage{url}            
\usepackage{nicefrac}       
\usepackage{microtype}      
\usepackage{stmaryrd}
\usepackage{graphicx}
\usepackage{booktabs}       
\usepackage{amsfonts,amsmath,amssymb}       
\usepackage{nicefrac}       
\usepackage{microtype}      
\usepackage[parfill]{parskip}
\usepackage{mathtools}
\usepackage{cases}
\usepackage{comment}
\usepackage{paralist}
\usepackage{caption}
\usepackage{algorithmic}
\usepackage{breqn}
\usepackage{algorithm}
\usepackage{color}
\usepackage{appendix}
\usepackage{xspace}
\usepackage{enumitem}
\usepackage{gensymb}
\usepackage[english]{babel}
\usepackage{xcolor}
\usepackage{wrapfig}
\usepackage{bbm}
\usepackage{paralist}
\usepackage[export]{adjustbox}
\usepackage[load-configurations=version-1]{siunitx} 

\theorembodyfont{\upshape}
\theoremheaderfont{\scshape}
\theorempostheader{:}
\theoremsep{\newline}

\newcommand{\St}{S}

\newcommand{\dimSt}{d_s}

\newcommand{\Ob}{O}

\newcommand{\Ac}{A}

\newcommand{\xmark}{\text{\ding{55}}}

\jmlrvolume{XX}
\jmlryear{2020}
\jmlrsubmitted{LEAVE UNSET}
\jmlrpublished{LEAVE UNSET}
\jmlrworkshop{Machine Learning for Health (ML4H) 2020} 

\title[Representation Learning for RL in Healthcare]{An Empirical Study of Representation Learning\\ for Reinforcement Learning in Healthcare}

\author{
\Name{Taylor W. Killian} \Email{twkillian@cs.toronto.edu}\\
\Name{Haoran Zhang} \Email{haoran@cs.toronto.edu}\\
\addr University of Toronto, Vector Institute
\AND 
\Name{Jayakumar Subramanian} \Email{jayakumar.subramanian@gmail.com}\\
\addr Media and Data Science Research Lab, Adobe India
\AND
\Name{Mehdi Fatemi} \Email{Mehdi.Fatemi@microsoft.com}\\
\addr Microsoft Research
\AND
\Name{Marzyeh Ghassemi} \Email{marzyeh@cs.toronto.edu}\\
\addr University of Toronto, Vector Institute
}

\begin{document}

\maketitle

\begin{abstract}
Reinforcement Learning (RL) has recently been applied to sequential estimation and prediction problems identifying and developing hypothetical treatment strategies for septic patients, with a particular focus on \emph{offline} learning with observational data. 
In practice, successful RL relies on informative latent states derived from sequential observations to develop optimal treatment strategies. 
To date, how best to construct such states in a healthcare setting is an open question.
In this paper, we perform an empirical study of several information encoding architectures using data from septic patients in the MIMIC-III dataset to form representations of a patient state.
We evaluate the impact of representation dimension, correlations with established acuity scores, and the treatment policies derived from them.
We find that sequentially formed state representations facilitate effective policy learning in batch settings, validating a more thoughtful approach to representation learning that remains faithful to the sequential and partial nature of healthcare data.

\end{abstract}

\section{Introduction}
\label{sec:intro}

\begin{figure}[ht!]
\begin{center}
\includegraphics[width=0.85\linewidth]{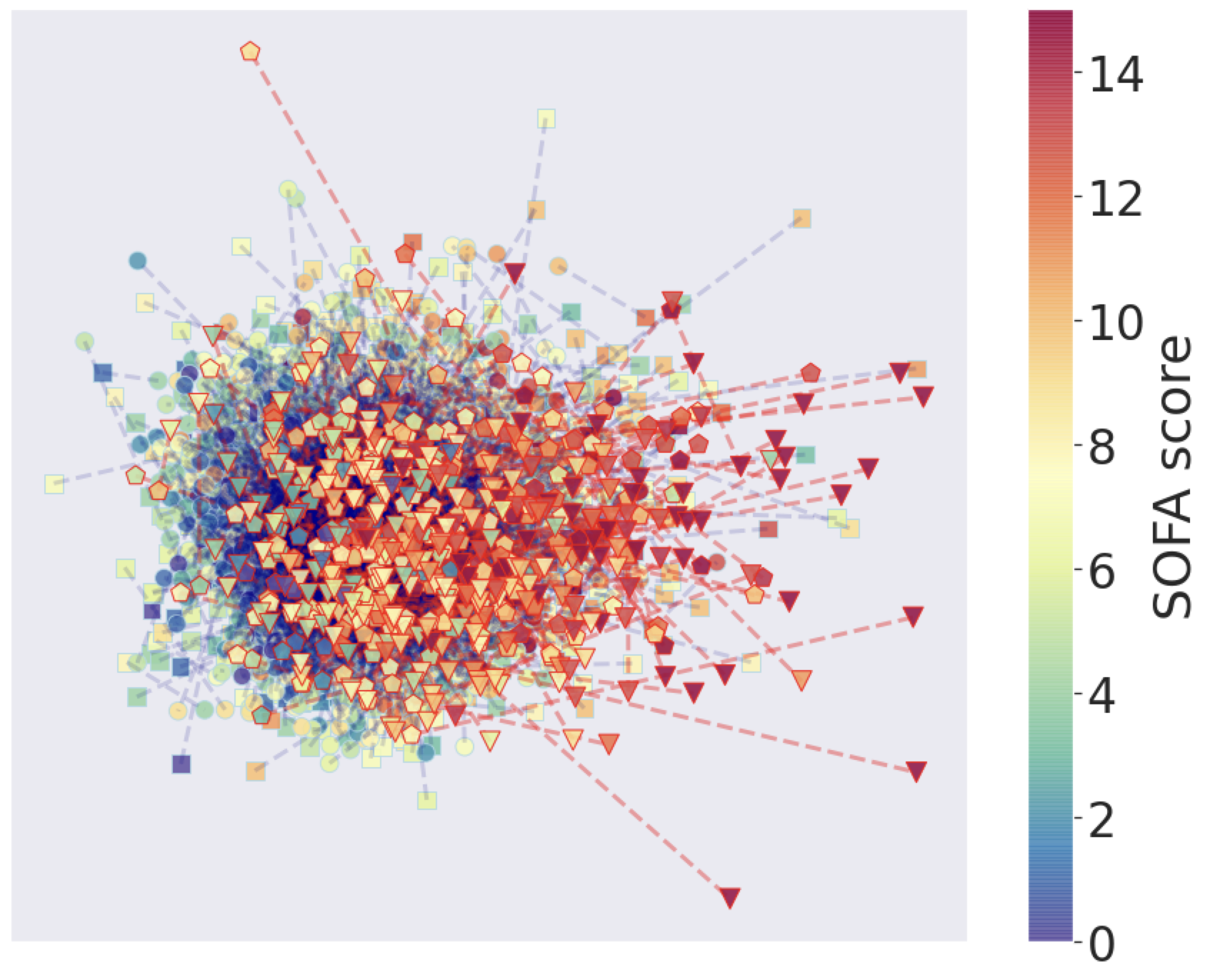}
\vspace{-0.25cm}
\caption{\footnotesize The first and final observations of septic patients in MIMIC-III, colored by SOFA score, visualized via Principal Component Analysis. Blue lines connect observations of patients who recovered, while red lines signify those that did not. Notably, these raw observations of severe health are not directly separable.}
\label{fig:intro_pca}
\end{center}
\vspace{-.35cm}
\end{figure}

Many problems in healthcare are a form of sequential decision making, e.g., clinical staff making decisions about the best ``next step'' in care \citep{ghassemi2019practical}. 
Solving these problems is similar to finding an optimal decision making policy, requiring estimation and optimization of the cumulative effects of decisions over time~\citep{sox2007medical}.
Recently, reinforcement learning (RL) has been proposed as a promising approach for finding an optimal policy for such processes from data~\citep{gottesman2019guidelines}.
However, the development of a successful policy rests on the ability to derive informative states from observations. In healthcare these observations are noisy, irregular, and may not convey the entirety of a patient's condition \citep{obermeyer2016predicting}. While there are many proposed state construction approaches to handle these challenges~\citep{li2019optimizing,chang2019dynamic,peng2018improving,prasad2017reinforcement,raghu2017deep,RaghuKomorowskiCeliEtAl2017}, few consider the sequential nature of observations, choosing instead to isolate the features from a single time step to construct the state.

Critical care is one specific setting where sequential data is crucial for predictive modelling. Raw physiological observations may not be clearly separable with respect to patient acuity or outcome, complicating downstream prediction and treatment models~\citep{ibrahim2020classifying} (see Figure~\ref{fig:intro_pca}). Complex model architectures have shown improved performance on such tasks, due in part to their improved ability to generate high-quality representations~\citep{choi2016multi,sadati2018representation,weng2019representation}. Yet within healthcare, the design and learning of patient representations for RL is an open problem~\citep{yu2019reinforcement}. 

In this work, we provide a controlled investigation of sequentially encoded state representations for use within RL applied to healthcare. We focus on the problem of treating septic patients~\citep{liu2020reinforcement}, using a patient cohort defined by~\citet{KomorowskiCeliBadawiEtAl2018} from the MIMIC-III dataset~\citep{mimicweb}.
We compare seven encoding architectures, and evaluate representations learned from sequential patient observations through three experiments. 

First, we examine the effect of representation dimension when training models to predict \underline{\bf s}ubsequent physiological \underline{\bf o}bservations (SO) through autoencoding~\citep{baldi2012autoencoders} prior physiological observations. We position this as an auxiliary task to the development of a treatment policy~\citep{jaderberg2017reinforcement}. 

Second, we investigate the impact of including contextual information as well as regularization when training these models. Context is added by augmenting the physiological observations with 5 demographic features. When regularizing model training, the learned representations are regularized to correlate with three clinical patient acuity scores -- OASIS \citep{jones2009sequential}, SAPS II \citep{le1993new} and SOFA \citep{johnson2013new}. We then qualitatively evaluate representations to determine their correlation with these scores, and embed them into a lower dimensional visualization to demonstrate their separability in contrast to the raw data.

Finally, we learn treatment policies from the encoded patient state representations using a state of the art off-policy RL algorithm, the discretized form of Batch Constrained Q-learning (dBCQ)~\citep{fujimoto2019benchmarking}. Policies are evaluated using weighted importance sampling~\citep{mahmood2014weighted}.

To our knowledge, we present the first rigorous empirical evaluation of learned patient state representations that facilitate policy learning. 
A summary of our contributions are:
\begin{itemize}[leftmargin=*]
\item We show that, keeping all other hyperparameters constant, increasing the latent dimensionality could reduce prediction accuracy, indicating that high capacity representations are not always most informative.
\item We find that including demographic context when learning the state representation generally improves the performance of predicting SO. 
\item We demonstrate that sequentially formed state representations can facilitate effective policy learning in batch settings. In particular, we find that representations learned through the recent Neural CDE~\citep{kidger2020neural} facilitate an especially effective policy.
\end{itemize}

\section{Background and related work}
\label{sec:background}
State representation learning has a long history within RL as a primary means of making complex control tasks computationally tractable~\citep{sutton1999between}.
Recent research has also separated feature extraction from policy learning~\citep{raffin2019decoupling}, where the goal is to isolate relevant features of the recorded observations in the representation, and provide more salient information to the policy learning algorithm. 

Problems modeled as POMDPs often require a state representation to be specified, typically deriving from prior observations and actions~\citep{kaelbling1998planning}. 
Past work in state construction has ranged from concatenation of a finite number of consecutive observations~\citep{mnih2013playing} to using the final layer of a recurrent neural network (RNN) to collectively embed a sequence of inputs~\citep{hausknecht2015deep}. 

Most prior work in the context of RL and healthcare has constructed states from unprocessed observations, framing the problem as a fully observable MDP~\footnote{For reference, Table~\ref{tab:background} summarizes these approaches, found in Appendix~\ref{sec:apdx_prior_table}}. This approach naively abstracts the true nature of the data generating process which is inherently partially observable. Missingness as well as an incomplete understanding of biological and physiological processes contribute to the partial nature of healthcare observations. There is a growing set of RL literature in this applied space that accounts for partial observability explicitly. The literature specific to sepsis treatment~\citep{tsoukalas2015data,li2018actor,peng2018improving, li2019optimizing,lu2020deep} often learns state representations by utilizing recurrent methods, encoding sequentially observed features of the patient's condition into a hidden state. 

To date, none of these works provide any analysis or justification of specific state representation choices. In this paper we address this empirical gap by rigorously evaluating multiple recurrent state representation learning approaches for use in healthcare. With this study we hope to provide a foundation for further research into representation learning for sequential decision problems within healthcare.

\section{Data}
\label{sec:dataset}
We consider the treatment of septic patients using data from the Medical Information Mart for Intensive Care (MIMIC-III) dataset (v1.4) \cite{johnson2016mimic}. 
We follow~\citet{KomorowskiCeliBadawiEtAl2018} to extract and preprocess\footnote{Code available at \url{https://github.com/matthieukomorowski/AI_Clinician}} relevant vital and lab measurements to build a cohort of 19,418 patients among which there is an observed mortality rate just above 9\% (determined by death within 48h of the final observation). 

To evaluate the formation of sequential representations of a patient's condition, we focus on patient vital signs and lab measurements that change over time, whether in response to selected treatments or as a consequence of their acute condition. This creates a dataset of $33$ features $\mathcal{O}$ with a discrete categorical action space with $25$ possible choices of combination between fluid and vasopressor amounts. We also experiment with including 5 additional demographic features $\mathfrak{D}$.

We include a list of features in Table~\ref{tab:features} with additional details included in Section~\ref{sec:apdx_data} of the Appendix.

\section{Methods}
\label{sec:methods}

\begin{figure*}[ht]
    \vspace{-0.5cm}
    \hspace{-.75cm}
    \includegraphics[width=1.15\textwidth]{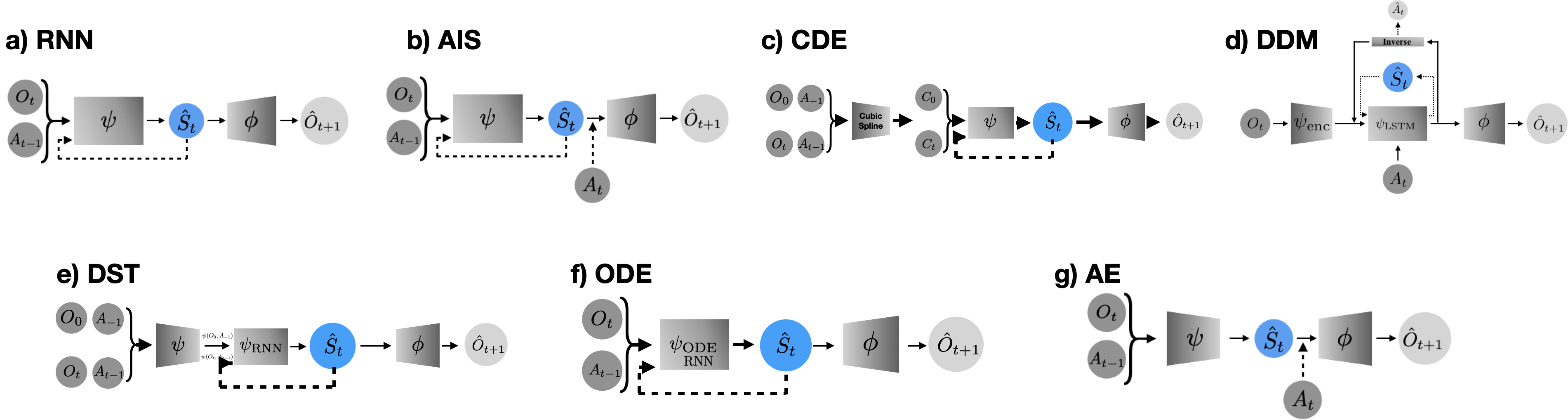}
    \caption{\footnotesize The architectures used to construct state representations via predicting future observations. {\bf a)} basic RNN autoencoder {\bf b)} Approximate Information State~\citep{AIS} {\bf c)} Neural CDE~\citep{kidger2020neural} {\bf d)} Decoupled Dynamics Module~\citep{zhang2018decoupling} {\bf e)} Deep Signature Transform~\citep{bonnier2019deep} {\bf f)} ODE-RNN~\citep{rubanova2019latent} {\bf g)} a non-recurrent Autoencoder. See Table~\ref{tab:arch_overview} in Appendix, Section~\ref{sec:apdx_arch} for a summary.}
    \label{fig:all_archs}
    \vspace{-0.5cm}
\end{figure*}

In this section, we provide a general overview of state representation learning via autoencoding architectures. We focus on the context of our first experimental analysis, where representations are used to predict the subsequent observation (SO). 

\subsection{General overview}
With a batch of observed patient trajectories---comprised of transitions between subsequent observations $\Ob_{t}$ and $\Ob_{t+1}$ following treatment action $\Ac_t$---we seek to learn an encoding function $\psi: \mathcal{H}_{t,t-1}\rightarrow\hat\St_t$ as well as a decoding prediction function $\phi:\hat\St_t\times\Ac_t\rightarrow\hat\Ob_{t+1}$. Here the history $\mathcal{H}_{t, t-1}$ contains all observations $\Ob_{0:t}$ and actions $\Ac_{0:t-1}$ preceding the target observation $\Ob_{t+1}$. Together, the encoding and decoding functions form a prediction $\hat\Ob_{t+1}$ using the learned state representation $\hat\St_t$. 
That is, {\small $\hat\Ob_{t+1} = \phi\left(\psi\left(\mathcal{H}_{t,t-1}\right), \Ac_t\right) = \phi(\hat\St_t, \Ac_t)$.} 
To facilitate sequentially stable predictions for the state representation $\hat\St_t$ we choose encoding functions $\psi$ with a recurrent structure. Thus, $\hat\St_t$ implicitly embeds the history $\mathcal{H}_{t,t-1}$, which has been shown to improve sepsis treatment policies~\citep{li2019optimizing}.

We jointly train the encoding function $\psi$ and decoding function $\phi$ via a loss function $\mathcal{L}(\Ob_{t+1}, \hat\Ob_{t+1})$, which in general computes the mean squared error between the predicted and true SO, as specified by the particular encoding approach.

\subsection{Information encoding models}
\label{sec:models}
We target six recurrent modeling approaches, largely motivated by their development to learn dynamics models: 
\begin{itemize}[leftmargin=*,noitemsep]
\item Basic RNN Autoencoder (RNN)~\citep{chung2014empirical}
\item Approximate Information State (AIS,~\cite{AIS})
\item Neural Controlled Differential Equations (CDE,~\cite{kidger2020neural})
\item Decoupled Dynamics Module (DDM,~\cite{zhang2018decoupling})
\item Deep Signature Transforms (DST,~\cite{bonnier2019deep})
\item And the ODE-RNN (ODE,~\cite{rubanova2019latent})
\end{itemize}
These approaches are depicted in Figure~\ref{fig:all_archs}. We also compare these approaches to a simple non-recurrent Autoencoder (AE). A comparative overview of the features that differentiate each approach as well as specific details about how each are trained are presented in Sec.~\ref{sec:apdx_arch} of the Appendix.

The unifying feature among these approaches is the development of a latent representation space $\hat\St$ that encodes information about the patient observations made over time. The formation of $\hat\St$ is meant to develop informative representations to facilitate better downstream policy learning, by implicitly accounting for the history $\mathcal{H}_{t, t-1}$. That is, we seek to develop a strategy to select treatments based on the encoded history via the learned state representation: $\Ac_t \sim \pi(\hat\St_t|\mathcal{H}_{t,t-1})$. Through the remainder of this work, we evaluate the characteristics of the representations embedded in $\hat\St$.

\noindent{\bf Model training:} We separate the data into a 70/15/15 train/validation/test split using stratified sampling. This maintains the same proportions of each terminal outcome (survival or mortality), and ensures that no patients are repeated across splits. All models were trained for the same number of epochs, using a variety of learning rates and 5 random initializations. The final settings for each model architecture are provided in the Appendix, Section~\ref{sec:apdx_arch}. 

\subsection{Augmenting the learning process}
\label{sec:augmentation}
In hopes of ensuring that the intermediate state representations $\hat\St_t$ retain clinically relevant features, we investigate augmenting the training of the representation space $\St$ through a combination of two options:  \begin{inparaenum} \item[(1)] Include the demographic context features $\mathfrak{D}$ (e.g. age, gender, etc.) as input to the encoder function $\psi$. When training with this option the history $\mathcal{H}_{t,t-1}$ contains observations $\Ob_i^+ = [\Ob_i, \mathfrak{D}_i]$. \item[(2)] Regularize the loss function by the Pearson correlation between the state representation and a set of acuity scores derived from the patient observations. \end{inparaenum} We utilize three independent acuity scores --- SOFA, SAPS II and OASIS --- through a linear combination of the correlation coefficients to subtract from the loss. The complete objective function when using this form of regularization is then,
\begin{equation*}
    Loss = \mathcal{L}(\Ob_{t+1}, \hat\Ob_{t+1})  -\lambda~\rho(\hat\St_t)
\end{equation*}
where {\small $\lambda~\rho(\hat\St_t) = \lambda_1~\rho^{\text{SOFA}}(\hat{\St_t}) + \lambda_2~\rho^{\text{SAPS II}}(\hat\St_t) + \lambda_3~\rho^{\text{OASIS}}(\hat\St_t)$}. We choose the hyperparameter $\lambda$ so that the final prediction loss of the regularized model is not inordinately larger than its unregularized counterpart. Additionally, we set $\lambda_1=\lambda_2=\lambda_3$ for simplicity in this paper yet these hyperparameters could be chosen independently of one another.

\subsection{Policy development}
\label{sec:policymethods}
We train policies on each of the learned state representations outlined in Section~\ref{sec:models}. As we do not have the ability to generate more data through an exploration of novel treatment strategies, we develop a policy using offline, batch reinforcement learning. In this setting, it is critical that the estimated value function not extrapolate to actions that are absent from the provided data~\citep{gottesman2019guidelines}. To avoid this extrapolation error \citet{fujimoto2019off} developed an algorithm that truncates any Q-function estimate corresponding to actions that fall outside the support of the dataset. This algorithm, Batch Constrained Q-Learning (BCQ), originally designed for continuous control problems was later adapted for use in discrete action settings~\citep{fujimoto2019benchmarking}.

We use this discretized BCQ algorithm to learn treatment policies from state representations $\hat\St$. We train the policies using the encoded training subset of our data and validate the performance with the testing subset using weighted importance sampling (WIS), following~\cite{li2019optimizing}. The WIS return for each policy throughout training is computed by: $R^{\mathrm{WIS}} = \frac{\sum_n^N w_n R_n}{\sum_n^N w_n}$, where the $w_n$ are the per-trajectory $\mathrm{IS}$ weights and $R_n$ is the observed outcome of the trajectory. All further details regarding policy training and intermediate results are provided in Section~\ref{sec:apdx_policy_training}.

\section{Empirical Study}
\label{sec:empirical_study}
We evaluate the representations $\hat S_t$ learned from patient data following the three experiments outlined at the conclusion of Section~\ref{sec:intro}. All analyses and results reported through the remainder of this section are provided using only the test set of the patient cohort. All code used to extract and preprocess the data, train and evaluate the encoding models as well as the policies can be found at \url{https://github.com/MLforHealth/rl_representations}.

\subsection{Representation dimension in SO prediction}
\label{sec:prediction}
We evaluate the accuracy of predicting the SO $\Ob_{t+1}$ from $\Ob_t$ and $\Ac_t$. Our primary investigation considers the effect of varying the dimension $\hat\dimSt$ of the learned state representation $\hat\St_t$ from the set {\small $\hat \dimSt\in\{4,8,16,32,64,128,256\}$}. Other than varying the latent dimension in each run, we keep all other model and optimization hyperparameters constant. This experiment evaluates the information capacity needed in the state representation $\hat\St_t$ to adequately predict the SO. 

\begin{figure}[ht]
  \hspace{-0.2cm}
  \includegraphics[width=1.05\linewidth]{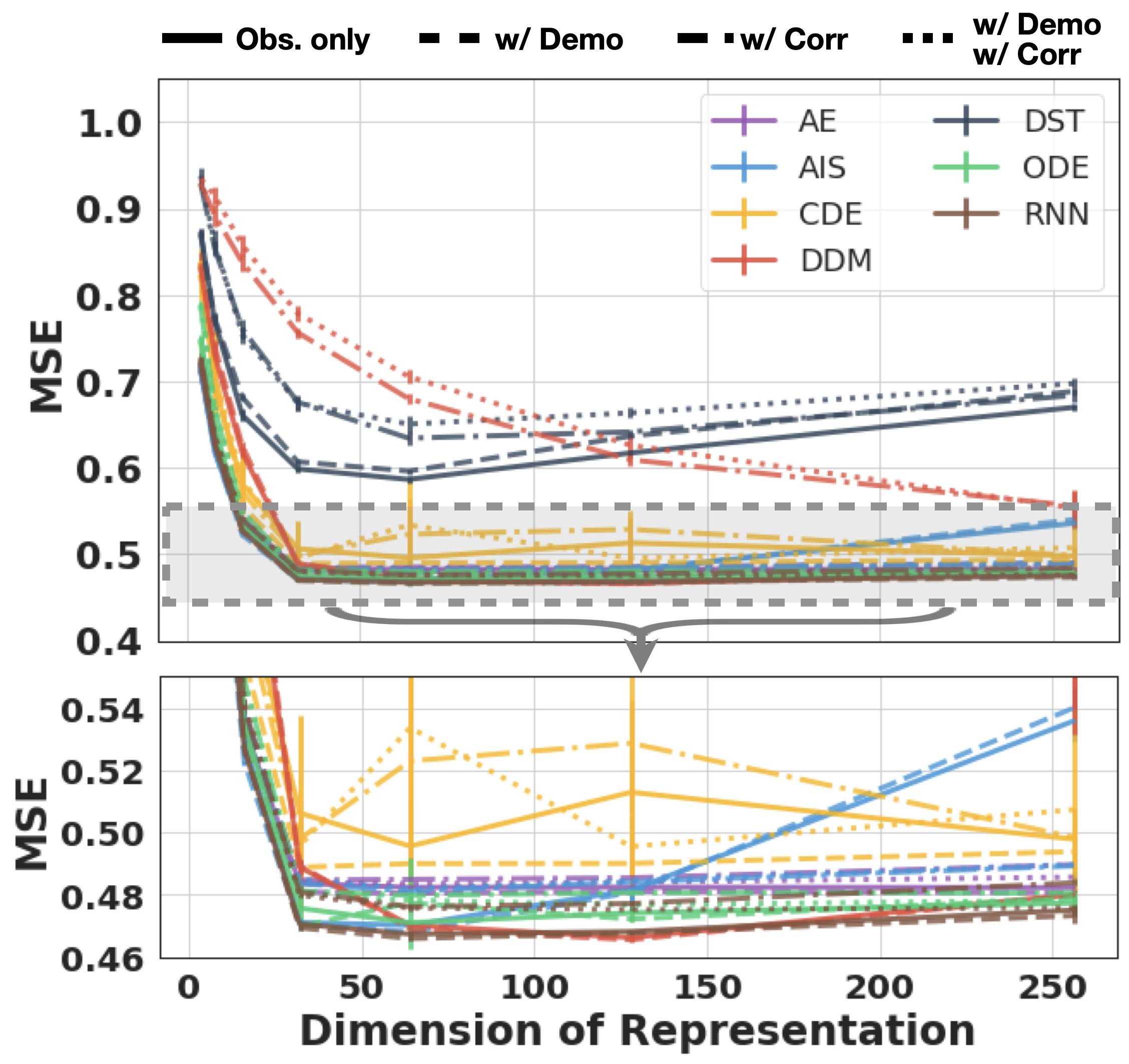}
  \caption{\footnotesize Mean squared error for SO prediction as a result of varying $\hat\dimSt$, comparing various training settings. Error bars are twice the std. dev. of each model over 5 random seeds. We note that augmenting the input to the encoding function $\psi$ with demographic context generally improves prediction performance. See Table~\ref{tab:next_step} for the best performing settings.}
  \label{fig:next_step_results}
  \vspace{-0.5cm}
\end{figure}

Results from these models, learned through the described training settings, are presented in Figure~\ref{fig:next_step_results} and Table~\ref{tab:next_step}. We see that the prediction performance of these models saturates as the dimension increases beyond 64, with the test loss increasing with larger representations. Aside from DST, the best performing settings of all other approaches converge between a loss of $0.46$ and $0.48$. This indicates that the highest capacity representations may not be the most informative for this prediction task.

\subsection{Augmenting learning in SO prediction}
We evaluate the two proposed training augmentations (see Sec.~\ref{sec:augmentation}) --- adding demographic features $\mathfrak{D}$ during training, and regularizing $\hat\St$ to be correlated with SOFA, SAPS II and OASIS --- via the accuracy of predicting the SO $\Ob_{t+1}$ from $\Ob_t$ and $\Ac_t$.

When augmenting the input to the encoding function $\psi$ with demographic context $\mathfrak{D}$ the prediction performance is generally improved (see the dashed curves in Figure~\ref{fig:next_step_results}). In contrast, the performance slightly degrades when the learned representations are regularized to be correlated with acuity scores (see the dotted and dot-dash lines in Figure~\ref{fig:next_step_results}), except for the DDM and DST models where there is a noticeable negative effect on model performance. 

\begin{table}[ht]
\centering
  \vspace{-0.15cm}
  \caption{\footnotesize Optimal model settings for each approach when predicting the SO. Models are trained with observations $\mathcal{O}$ and can be augmented with demographic context $\mathcal{D}$ or by the correlation regularization $\mathcal{C}$.}
  \label{tab:next_step}
  \vspace{-0.25cm}
  {\small
  \resizebox{\linewidth}{!}{
  \begin{tabular}{cccc}
    {\bf Approach} & {\bf Best MSE} & {\bf $\hat \dimSt$} & {\bf Training Setting}\\ \midrule\midrule
    AE & 0.4804$\pm$0.001 & 64 & w/ $\mathcal{O} + \mathfrak{D}$ \\
    \midrule
    AIS & 0.4679$\pm$ 0.004 & 64 &$\mathcal{O} + \mathfrak{D}$  \\
    CDE &  0.4887$\pm$ 0.019 & 32 & $\mathcal{O} + \mathfrak{D}$  \\
    DDM & 0.4654$\pm$ 0.002 & 128 & $\mathcal{O} + \mathfrak{D}$  \\
    DST & 0.5863$\pm$ 0.013 & 64 & $\mathcal{O}$ \\
    ODE & 0.4698$\pm$ 0.003 & 32 & $\mathcal{O} + \mathfrak{D}$  \\
    RNN & 0.4658$\pm$ 0.002 & 64 & $\mathcal{O} + \mathfrak{D}$ \\
  \bottomrule
\end{tabular}
}
}
\end{table}

\subsection{Qualitative analysis}

The following analyses investigate the qualitative impact that the separate training strategies have on learned representations. 

\noindent{\bf Representation-to-acuity score correlation} We first evaluate the average correlation coefficient between the representations and derived acuity scores (see Section~\ref{sec:apdx_acuity} for more information). This is to demonstrate the capacity of the representations $\hat\St_t$ to maintain clinically relevant information. We perform this analysis with and without the correlation regularization described in the previous subsection. The intention of this regularization is that the more positively correlated the representation is to the acuity scores, the more clinically informative the learned representation is. This was designed in hopes to improve SO prediction and policy learning yet there was no demonstrated advantage in doing so as shown in Figure~\ref{fig:next_step_results} and Figure~\ref{fig:apdx_policy_training}.

\begin{figure}[ht]
\centering
\includegraphics[width=\linewidth]{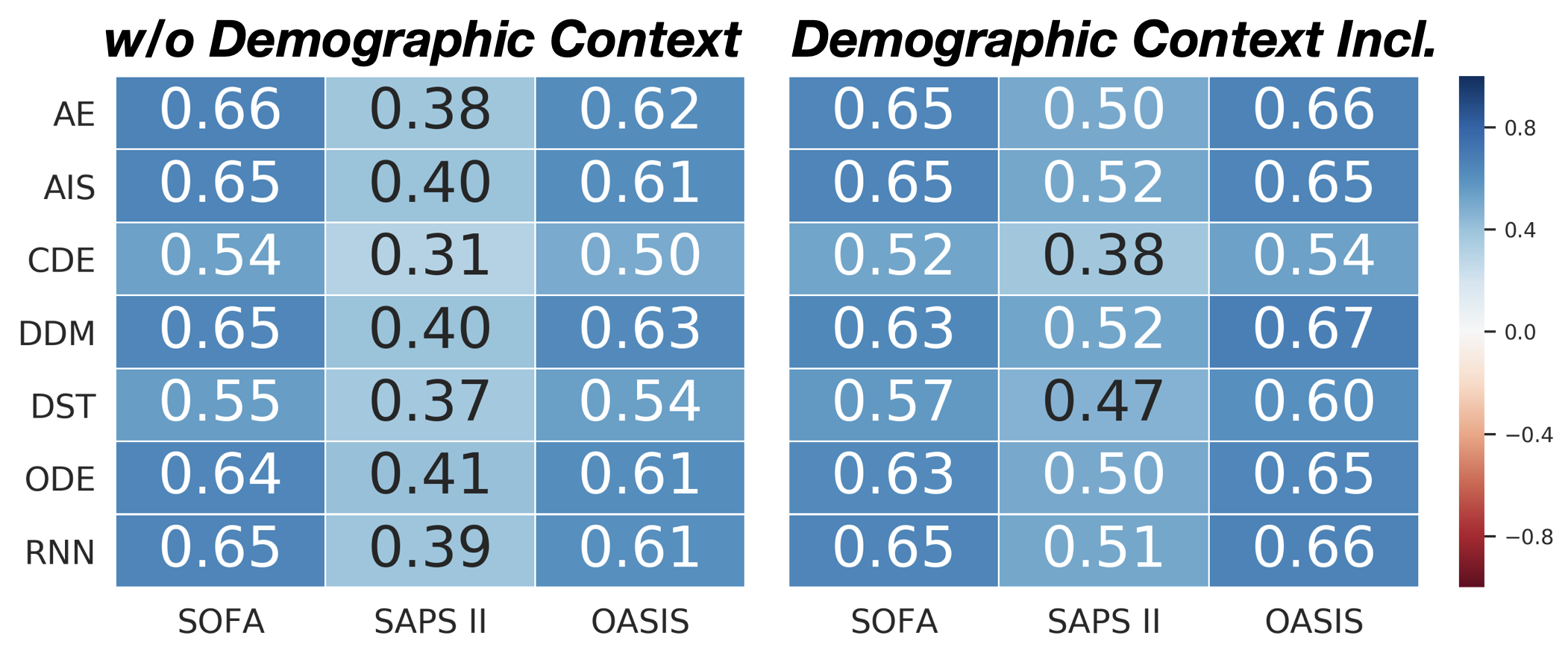}  
\caption{\footnotesize The average Pearson correlation coefficient between the state representations from each encoding approach and acuity scores. Shown here are the average coefficients when regularizing the learning process and demographic features are omitted (left) or are included (right) as input. For SAPS II and OASIS, the inclusion of demographic features when constructing the state representations results in higher correlation.}
\label{fig:corr_heatmap}
\end{figure}

We show the average correlation coefficients of the learned state representations with acuity scores in Figure~\ref{fig:corr_heatmap} for the two training settings where regularization is included (with and without demographic context). Unregularized representations fail to encode information that is correlated with the acuity scores (see Figure~\ref{fig:apdx_corrHtmp} in the Appendix). Between the two settings, representations are better correlated with the acuity scores when a patient's demographic context is included. This suggests that clinical acuity scores are strongly entangled with demographics features. Further investigation into the effects of this entanglement, including questions of fairness, is outside the scope of this study and is therefore a suggested element of future work.

\begin{figure*}[ht!]
\centering
\includegraphics[width=\linewidth]{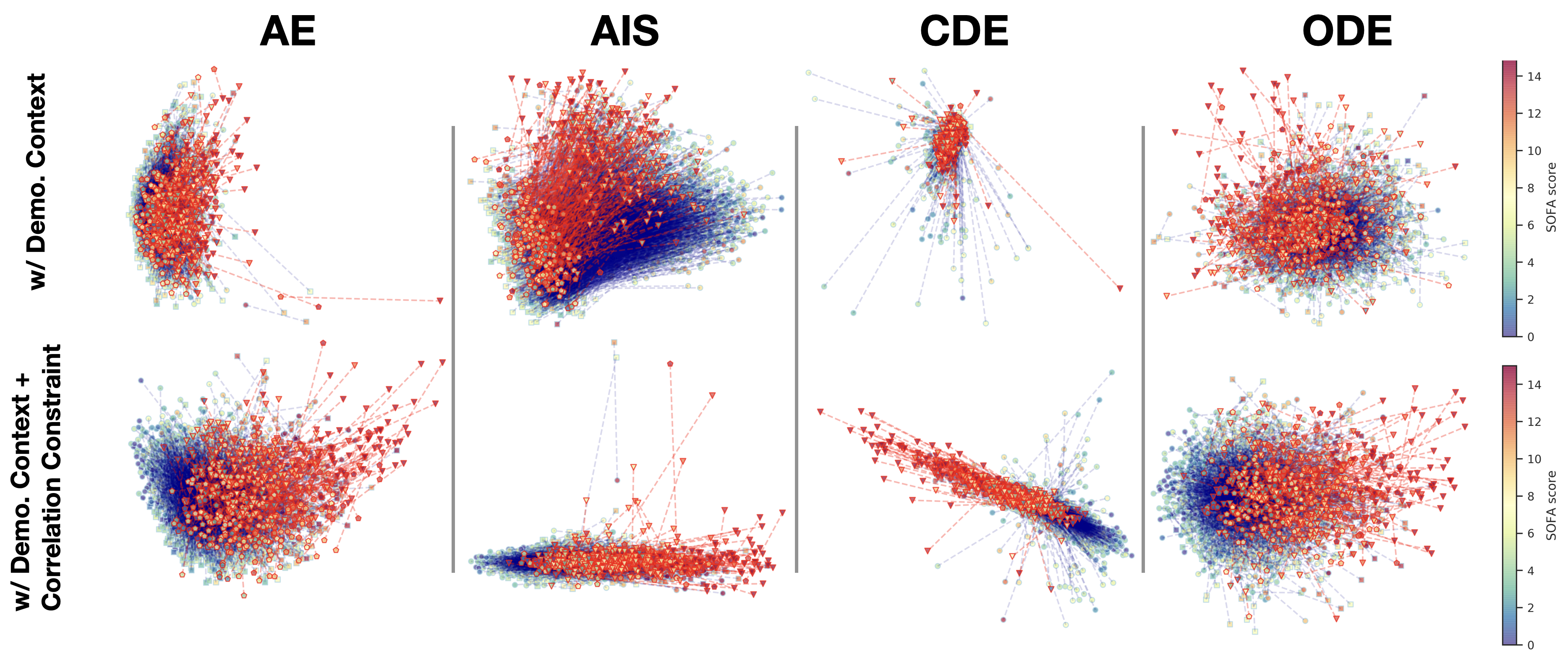}
\caption{\footnotesize Representations of patient health, learned through a non-recurrent autoencoder (AE), Approximate Information State (AIS), a Neural CDE (CDE) and an ODE-RNN (ODE) (left-to-right, all other approaches are included in the Appendix, Section~\ref{sec:apdx_pca}) for two training settings. We show the first and final observations made of septic patients in the MIMIC-III dataset, colored by the SOFA score. Blue lines represent the trajectory of patients who recovered, while red lines connect observations of those those that did not.}
\label{fig:comb_pca}
\vspace{-0.5cm}
\end{figure*}

\noindent{\bf Visualizing learned state representations} Next, we use principal component analysis (PCA) to project the learned representations into a lower dimensional space. PCA embeddings are fit using the encoded representations for the entire test set but only the first and final representation from a patient trajectory are vizualized. To aid in connecting these two points, we have drawn a line between them colored by the patient outcome, survival (blue) vs. death (red).

As shown in Figure~\ref{fig:intro_pca}, PCA projections of raw observations are not separable. Separability is desirable because a representation that separates patients who are most at risk of death could be used to more easily facilitate prediction models. 
In Figure~\ref{fig:comb_pca} we show PCA projections for AE, AIS, CDE and ODE in two training settings (remaining approaches and training settings in the Appendix, Section~\ref{sec:apdx_pca}). We focus on the role of including demographics without acuity regularization (top), and when it is included (bottom). With exception of AIS, regularization provides better separation between the patients that survive their sepsis infection and those that do not. Additionally, the regularization compresses the feature space of some encoding approaches. In combination with findings in Section~\ref{sec:prediction}, this compression suggests that the information prioritized via acuity regularization does not contribute to an improved state representations despite improved separability in representation space. Further analysis of the information content stored in the representations as a consequence of being regularized to correlate with acuity scores is a subject of future work.

\subsection{Policy training and evaluation} 
\label{sec:policy_training}
We investigate the quality of treatment policies learned from the state representations via the approaches outlined in Section~\ref{sec:methods}, following the procedure outlined in \ref{sec:policymethods}. We train policies using discretized BCQ, and evaluate with weighted importance sampling (WIS).

\begin{figure}[ht]
  \centering
  \includegraphics[width=\linewidth]{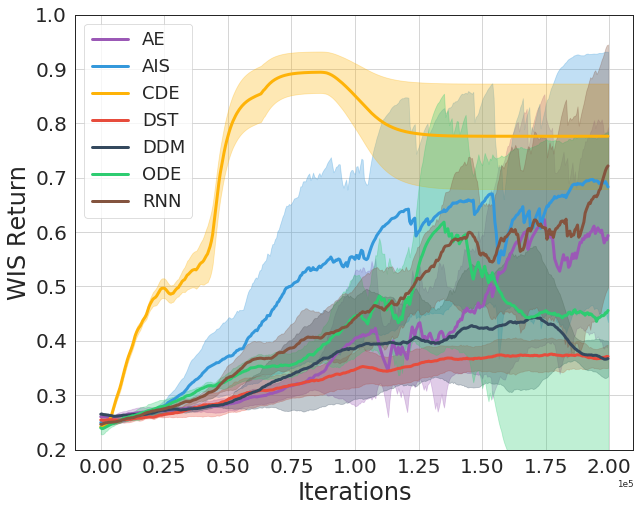}
  \caption{\footnotesize WIS evaluation of policies trained from the representations encoded by the architectures outlined in Section~\ref{sec:methods}. Policies are trained from an experience replay buffer comprised of the training batch of patient trajectories for 200k iterations, evaluating the trained policy every 500 iterations. Results presented here are averaged over 5 random seeds, the shaded region measures a single standard deviation across seeds.}
  \label{fig:policy_comparison}
  \vspace{-0.35cm}
\end{figure}

In Figure~\ref{fig:policy_comparison} we present the best performing policies learned from state representations. For each approach, excepting ODE, the top policies were learned from representations trained with the demographic context included as input to the encoding function $\psi$. The best ODE policy was developed from representations learned from the observations alone (see Figure~\ref{fig:apdx_policy_training} in the Appendix). 

Among the various approaches, policies learned from representations encoded by the Neural CDE (CDE) far outperform the others. Simpler recurrent based architectures such as AIS and RNN also obtain higher performance than the non-recurrent autoencoding baseline (AE). These results contribute toward the validation of our empirical hypothesis, that recurrent architectures provide better state representations in sequential partially observed settings. However, the AE based policy still learns a better policy than those based from far more complex methods (DST, DDM and ODE) signifying that the representations from these methods did not adequately encode sufficient information to learn a policy from in the batch setting, possibly due to dataset limitations.

\section{Discussion}
\label{sec:discussion}

In this paper we have empirically evaluated seven information encoding approaches to develop sequential state representations of patient health, useful for learning effective treatment policies. We performed several experiments to determine characteristics useful for training state representations from noisy patient data that is inherently partially observed. To support the formation of informative representations we designed a supervised task where the representation implicitly encodes a history $\mathcal{H}_{t,t-1}$ of previous observations and actions to predict the next SO. This auxiliary task allowed us to investigate several properties of the representation space $\hat\St$ based on decisions of how to execute the training. 

In Section~\ref{sec:prediction} we showed that higher dimensional representations reduce prediction accuracy, indicating the high capacity representations are not the most informative. In tandem we demonstrated that the inclusion of demographic context improves the learned state representations. This was verified (see Section~\ref{sec:apdx_policy_training}) when learning treatment policies. The best performing policies for each information encoding approaches presented in this paper were trained from representations learned with demographic context.

{\bf Future work} In future work we intend to explore the use of multi-task learning~\citep{mcdermott2020comprehensive,lin2019adaptive} to jointly train the representation space. Additionally, we plan to investigate methods that incorporate indicators of feature missingness and other underlying contextual variables~\citep{agor2019value,fleming2019missingness,sharafoddini2019new,che2018recurrent,lipton2016directly}. We intend to study the effect these approaches have on the representation space, including a quantification of any performance reductions that may arise through use of demographic information encoding bias in the representations~\citep{chen2018my}. 

The class of Neural Differential Equation methods~\citep{chen2018neural,rubanova2019latent,kidger2020neural} were developed to account for irregular time series with missing values and have demonstrated high performance in prediction tasks when provided feature sets with varying rates of missingness. Following the analyses performed in this paper, the Neural CDE appears promising for constructing state representations in the midst of the missingness and other irregularities inherent in healthcare data. 

The conceptual separation between representation learning and policy learning in this paper was motivated by prior literature on state representation learning~\citep{raffin2019decoupling}. This choice allowed us to focus on the formation and analysis of the representation space $\hat\St$. Another reason for this design choice was to enable straightforward use of current state of the art batch RL training algorithms. However, this decoupling is not necessary for developing off-policy sepsis treatment policies as discussed and demonstrated by~\cite{li2020optimizing}. Another line of future work utilizing the findings of this paper is to similarly develop an end-to-end policy development approach that combines the objectives of the auxiliary tasks and RL algorithm, explicitly accounting for the state representation space as it encodes features of the expected outcome via the RL objective.

Additionally, it is necessary to more fully evaluate and interpret what the learned state representations encode and whether clinically relevant relationships are preserved~\citep{bai2018interpretable}. It will be beneficial for the future use of these state representations to determine whether they embed trends in the data following the improving (or degrading) health of the patient beside only encoding features relevant for inferring the SO.

{\bf Conclusion} Such investigations and state representation learning will provide mechanisms by which we can better understand the cumulative effects of prescribed actions, chosen by following observed or learned policies.
State representations and learned value functions used in this manner can enable the identification of reliable treatment policies, developed following a learning process that acknowledges the sequential and partial nature of the observations that are made.

This paper recommends possible ways of thinking of representation learning as a form of auxiliary task within policy development. Among the various research directions that are natural extensions from this work, we affirm the necessity of thoughtfully designing the representation learning process to honor the partial and sequential nature of the data generating process. These opportunities for learning optimal state representations for RL in healthcare offer an exciting new area of research that we anticipate being fruitful for establishing future advances in clinically relevant sequential decision making problems.

\acks{\small We thank our many colleagues who contributed to thoughtful discussions and provided timely advice to improve this work. We specifically appreciate the feedback provided by Nathan Ng, Vinith Suriyakumar and Karsten Roth.

This research was supported in part by Microsoft Research, a CIFAR AI Chair at the Vector Institute, a Canada Research Council Chair, and an NSERC Discovery Grant.

Resources used in preparing this research were provided, in part, by the Province of Ontario, the Government of Canada through CIFAR, and companies sponsoring the Vector Institute \url{www.vectorinstitute.ai/\#partners}.}

\bibliography{bibliography}
\newpage
\appendix

\section{Details about Patient Cohort}
\label{sec:apdx_data}

\subsection{Data extraction and preprocessing}
To construct our patient cohort from the MIMIC-III database,we follow the approach described by~\citet{KomorowskiCeliBadawiEtAl2018} and the associated code repository given in~\cite{aiclinician}.
This includes all adult patients (aged 18 years and older) in the intensive care fulfilling the sepsis 3 criteria. A presumed onset of sepsis is defined by temporally related prescription of antibiotics and test results from microbiological cultures.
All patient observations are extracted in a 72h span around this presumed onset of sepsis (24h before presumed onset to 48h afterwards).
The original cohort extracted by~\citet{KomorowskiCeliBadawiEtAl2018} contained a set of 48 variables including demographics, Elixhauser status, vital signs, laboratory values, fluids and vasopressors received and fluid balance.
Missing or irregularly sampled data was filled using a time-limited sample-and-hold approach based on clinically relevant periods for each feature.
All values that remained missing after this step were imputed using a nearest-neighbor approach.
After imputation, all features are z-normalized.

Observed actions (administration of fluids or vasopressors) are categorized by volume and put into 5 discrete bins per action type. 
The combination of the type of actions leads to 25 possible discrete actions.

\subsection{Features used in this paper}

As described in Section~\ref{sec:dataset}, we only maintain features that correspond to continuous quantities, the evolution of which may result from the selected actions.
Those columns we remove from the original extracted cohort by~\citeauthor{KomorowskiCeliBadawiEtAl2018} are intended to be added to the learned state representations used for developing treatment policies.
We include the patient features used in this paper in Table~\ref{tab:features}.

\begin{table*}[ht]
\caption{\small Observed features used for learning state representations}
\label{tab:features}
\begin{center}
{\footnotesize
\begin{tabular}{|l|l|l|}
\multicolumn{3}{l}{\underline{Time-varying continuous features}}\\
\midrule
Glascow Coma Scale & Heart Rate & Sys. BP \\
Dia. BP & Mean BP & Respiratory Rate \\
Body Temp (C) & FiO2  &Potassium \\
Sodium & Chloride & Glucose \\
INR & Magnesium & Calcium \\
Hemoglobin & White Blood Cells & Platelets \\
PTT & PT & Arterial pH \\
Lactate & PaO2 & PaCO2\\
PaO2 / FiO2  & Bicarbonate (HCO3)  & SpO2 \\
BUN & Creatinine & SGOT \\
SGPT & Bilirubin & Base Excess\\
\midrule
\multicolumn{3}{l}{\underline{Demographic and contextual features}}\\
\midrule
Age & Gender & Weight \\
Ventilation Status & Re-admission status & \\
\bottomrule
\end{tabular}
}
\end{center}
\end{table*}

\subsection{Acuity Scores}
\label{sec:apdx_acuity}

Patient acuity scores are used in clinical practice to estimate the severity a patient's illness, and have historically been used as a predictor of mortality~\citep{silva2012predicting}. In order to constrain the learning of state representations we extract three acuity scores computed from the full patient observations from each 4h time step~\citep{hug2009icu}: Sepsis-related Organ Failure Assessment (SOFA)~\citep{vincent1996sofa}, Simplified Acute Physiology Score II (SAPS II)~\citep{SAPSII} and Oxford Acute Severity of Illness Score (OASIS)~\citep{johnson2017real}. For the particular heuristics used to calculate these scores, we refer the reader to the originating literature sources.

\subsubsection{Sepsis-related Organ Failure Assessment - SOFA}

The Sepsis-related Organ Failure Assessment score was developed to provide clinicians with an objective measure of organ dysfunction in a patient. The score is evaluated for 6 organ systems: pulmonary, renal, hepatic, cardiovascular, haematologic and neurologic. 
Under the Sepsis-3 criteria, a patient is presumed to be septic if the SOFA score increases by 2 or more points.

\subsubsection{Simplified Acute Physiology Score II - SAPS II}

The Simplified Acute Physiology Score II (SAPS II) was developed to improve issues with SAPS, a simplified score using 13 physiological parameters. 
These parameters were chosen using univariate feature selection to exclude features uncorrelated with hospital mortality.

\subsubsection{Oxford Acute Severity of Illness Score - OASIS}

The Oxford Acute Severity of Illness Score (OASIS) is a severity score developed algorithmically which directly optimized for clinical relevance, simultaneously performing multivariate feature selection. 
OASIS requires only 10 features, without depending on laboratory measurements, diagnosis or comorbidity information.

\section{Architecture Details}
\label{sec:apdx_arch}

We provide a comparative overview of the features that differentiate each approach in Table~\ref{tab:arch_overview}. Specific details about each architecture and how they are trained is included in the following subsections.

\begin{table*}[h!t]
\caption{\small Overview of approaches for state representation learning under evaluation}
\centering
\label{tab:arch_overview}
{\footnotesize
\begin{tabular}{ccccc}
\toprule
Approach & Recurrent &  Sequence as input & Num. Parameters \\ \midrule
AE  & & & $27k-76k$ \\ \midrule
AIS & $\xmark$ &  & $28k-339k$ \\ \midrule
CDE & $\xmark$ & $\xmark$ & $78.9k - 1.78m$ \\ \midrule
DDM & $\xmark$ &  & $6k-1.25m$ \\ \midrule
DST & $\xmark$ & $\xmark$ & $47k-256k$ \\ \midrule
ODE & $\xmark$ &  & $48.3k-329k$ \\ \midrule
RNN  & $\xmark$ & &  $26k-337k$  \\ \bottomrule
\end{tabular}
}
\end{table*}

\subsection{RNN}
\label{sec:apdx_archRNN}

\begin{figure}
\centering
\includegraphics[width=0.85\linewidth]{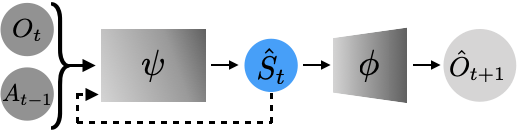}
\caption{\footnotesize A basic RNN architecture for SO prediction}
\label{fig:rnn_arch}
\end{figure}

Recurrent Neural Networks (RNNs) are extensions of conventional feed-forward neural networks capable of receiving correlated sequences as input. 
The RNN handles variable-length sequences by utilizing a recurrent hidden state, activated by features propagated from the previous timestep. 
When provided an observation $\Ob_t$ from a sequence, the RNN updates its recurrent hidden state $h_t$ by a nonlinear function that associates the $\Ob_t$ with $h_{t-1}$. 
Initially, this hidden state is set to a vector of zeros.
This hidden state, an embedding of the prior sequence of observations, can then be used to make predictions of various kinds depending on the specific context the model is trained for.
See \citep{chung2014empirical,jozefowicz2015empirical} for a more detailed introduction to such networks.

For predicting the SO in healthcare settings we make the following adjustments to a basic RNN architecture, shown in Figure~\ref{fig:rnn_arch}.
The current observation $\Ob_t$ is concatenated with the selected action and passed into the RNN along with the hidden state representation from the previous time step $\hat\St_{t-1}$. 
The hidden state representation $\hat\St_t$ is then passed to a decoder function $\phi$ that provides the prediction of the SO $\hat\Ob_{t+1}$.

We use a 3-layer Recurrent Neural Network (RNN) for estimating the encoding function $\psi$, where the first layer is a fully connected layer that maps the current observation and action (58 dimensional input: 33 dimensional observation with a 25 dimensional one-hot encoded action) to 64 neurons with ReLU activation.
This is followed by another $(64,128)$ fully connected layer with ReLU activation which is followed by a GRU layer~\citep{cho2014properties} with hidden state size $\hat\dimSt$ chosen from $\{4,8,16,32,64,128,256\}$.
For estimating the decoder function $\phi$, we use a 3-layer feed-forward neural network with sizes $(\hat\dimSt,64)$, $(64,128)$ and $(128,33)$ with ReLU activation for the first two layers.
The last layer outputs a 33-dimensional vector, which forms the mean-vector of a unit-variance multi-variate Gaussian distribution which is then used to predict the SO.

The best RNN architectures for each choice of $\hat\dimSt$ were trained for 600 epochs with a learning rate of $1e-4$. The $\lambda$s for regularizing the training to correlate with acuity scores are all set to 100.

\subsection{AIS}
\label{sec:apdx_archAIS}
The Approximate Information State (AIS)~\citep{AIS} was introduced as an approach to learning the state representation for POMDPs for use in dynamic programming.
The learned representation is defined in terms of properties that can be estimated from data, so it lends itself to be used in model pipelines where the state is used for some downstream task.
The function $\psi$ is comprised of an encoder followed by a gated recurrent unit~\citep{cho2014properties} which outputs the representation $\hat\St_t$. 
The input to $\psi$ is the concatenation of the observation $\Ob_t$ and last selected action $A_{t-1}$. 
The current action $\Ac_t$ (which is typically induced from the policy, conditioned on $\hat\St_t$) is concatenated to the state representation $\hat\St_t$ and then fed through the decoder function $\phi$ to predict the SO $\hat\Ob_{t+1}$.

\begin{figure}
\centering
\includegraphics[width=0.95\linewidth]{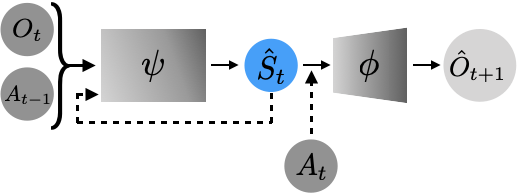}  
\caption{\footnotesize AIS architecture, adapted from ~\cite{AIS}}
\label{fig:ais_arch}
\end{figure}

AIS uses the same base architecture as the basic RNN with one adjustment.
For the decoder function $\phi$ we augment the input space by appending the current action $\Ac_t$ to the state representation $\St_t$. 
Therefore the AIS decoder function $\phi$, constitutes a 3-layer feed-forward neural network with sizes $(\hat\dimSt+25,64)$, $(64, 128)$ and $(128, 33)$ with ReLU activation for the first two layers. 
The last layer outputs a 33-dimensional vector, which forms the mean-vector of a unit-variance multi-variate Gaussian distribution which is then used to predict the SO.

The best AIS architectures for each choice of $\hat\dimSt$ were trained for 600 epochs with a learning rate of $5e-4$. The $\lambda$s for regularizing the training to correlate with acuity scores are all set to 100.

\subsection{DDM}
\label{sec:apdx_archDDM}

\citet{zhang2018decoupling}, introduced an model-based RL algorithm that decoupled dynamics and reward learning. This decoupling aimed to improve the generalization and stability of RL algorithms operating in environments where perturbations to the observations may occur. The dynamics module utilizes recurrent models to associate sequences of prior observations and their affect on subsequent observations.

\begin{figure}
\centering
\includegraphics[width=.95\linewidth]{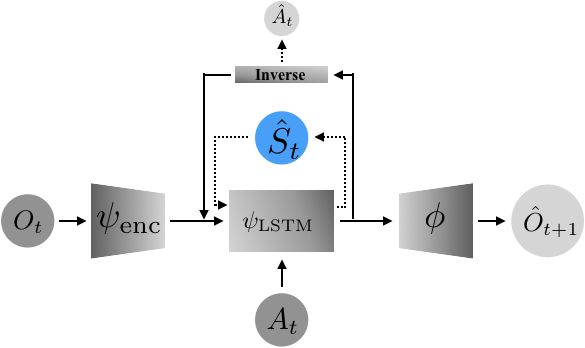}
\caption{The Decoupled Dynamics Module from~\citet{zhang2018decoupling}, adapted from its original presentation}
\label{fig:ddm_arch}
\end{figure}

We adapt this module, shown in Figure~\ref{fig:ddm_arch}, for the purpose of predicting the SO in a healthcare setting. 
The observation $O_{t}$ is provided to an encoder $\psi_{\text{enc}}$ the output of which is concatenated to the selected action $A_{t}$ and fed into an LSTM ($\psi_{\text{LSTM}}$)~\citep{hochreiter1997long} which provides the state representation $\hat{S}_{t}$. 
This state representation is then provided to the decoder function $\phi$ to provide a prediction of the SO $\hat{O}_{t+1}$. 
To stabilize the development of this learned state representation, $\hat\St_{t}$ is also fed to an inverse dynamics function (denoted by "Inverse" in Fig.~\ref{fig:ddm_arch}) along with the true SO to predict the action used to generate $\hat\St_{t}$.

For specific details about the set-up and training of decoupled dynamics module (DDM), we refer the reader to~\citet{zhang2018decoupling}\footnote{The author's code can be accessed at \url{https://github.com/facebookresearch/ddr}}.

The DDM archtecture is made up of three modules, an encoder ($\psi_{\text{enc}})$, a dynamics module ($\psi_{\text{LSTM}}$), and a decoder ($\phi$). 
These three modules combine to both create a latent embedding space for the state representations $\hat\St$ while also decoding these representations to predict the SO.
The encoding function $\psi_{\text{enc}}$ is comprised of a 3-layer feed-forward neural network with sizes $(33,\hat\dimSt)$, $(\hat\dimSt,288)$, $(288,\hat\dimSt)$. 
The first two layers are followed by exponential linear unit (ELU) activation functions. 
The final layer is passed through a \texttt{tanh} activation and provided as output to the dynamics model $\psi_{LSTM}$.

The dynamics module $\psi_{LSTM}$ receives as input the encoded observation  and SO, $z_t = \psi_{\text{enc}}(\Ob_{t})$, $z_{t+1} = \psi_{\text{enc}}(\Ob_{t+1})$ respectively the current action $\Ac_t$ and two separate hidden state vectors that describe the distribution of the latent distribution $\hat Z$ that the encoder produces estimates of with each observation. 
The dynamics module $\psi_{\text{LSTM}}$ begins with two linear layers of sizes $(25,\hat\dimSt)$ and $(\hat\dimSt, \hat\dimSt)$, the first of which has an ELU activation function. 
These layers embed the action $\Ac_t$. This embedding is concatenated with the encoded observation $z_t$ and passed through a linear layer with shape $(2*\hat\dimSt,\hat\dimSt)$. 
The output of this embedding is then passed to a LSTM Cell with input dimensions of dimension $\hat\dimSt$ and produces the mean and variance vectors of the latent distribution, each of size $\hat\dimSt$. 
The mean vector is then passed through a \texttt{tanh} activation function and provided as an estimate of the encoded SO $\hat{z}_{t+1}$. 
Finally, the dynamics module infers the action $\Ac_t$ that caused the transition between the encoded $z_t$ and $z_{t+1}$. 
These encoded representations of the observations are concatenated and passed through a 2-layer fully connected neural network, the first layer with shape $(2*\hat\dimSt, \hat\dimSt)$ followed by an ELU activation with the second layer having shape $(\hat\dimSt,25)$.

The decoder function $\phi$ is a 3-layer fully connected neural network. The first two layers have the shapes $(\hat\dimSt,288)$, $(288,\hat\dimSt)$ each followed by ELU activation functions. The final layer has the shape $(\hat\dimSt,33)$.
The decoder $\phi$ takes the predicted subsequent encoded observation ($\hat{z}_{t+1}$, which we use as our learned state representation) as input. 
The function outputs a 33-dimensional vector which is the prediction for the SO $\hat\Ob_{t+1}$.

The best DDM architectures were trained for 600 epochs with the following learning rates for each choice of $\hat\dimSt$;\\$\{4: 1e-3, \ 8:~1e-4, \ 16:~1e-4, \ 32:~5e-4, \ 64:~1e-4, \ 128:~1e-4, \ 256:~1e-4\}$ The $\lambda$s for regularizing the training to correlate with acuity scores are all set to $0.25$.

\subsection{DST}
\label{sec:apdx_archDST}
As outlined by~\citet{bonnier2019deep}, sequentially ordered data can have path-like structure. 
The statistics of such a path can be represented by the \textit{signature}~\citep{chevyrev2016primer}.
The mapping between a path and its signature is known as the signature transform. 
Neural network architectures that utilize such transforms may be capable of adequately handling irregularly sampled time-series data from partially observable environments such as those in healthcare.

\begin{figure}
\centering
\includegraphics[width=\linewidth]{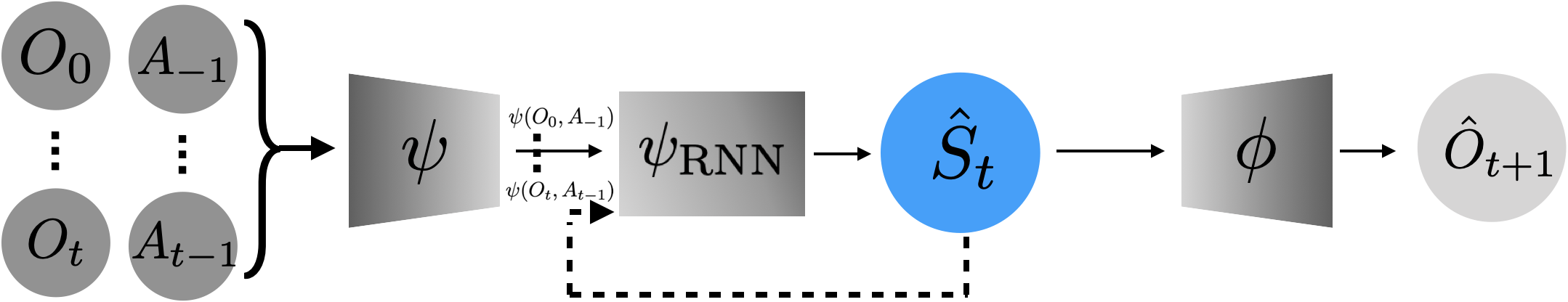}
\caption{\footnotesize The Deep Signature Transform architecture for SO prediction}
\label{fig:dst_arch}
\end{figure}

The signature transform $\text{Sig}^N$ is defined by an infinite sequence where $N$ roughly corresponds to the order of approximation of matching moments of a distribution. 
In practice, $\text{Sig}^N$ is truncated to include a finite number of elements.
The choice of $N$ and the dimension $d$ of the data points of the sequence influence the subsequent number of terms in the truncated signature as {\small $|\text{Sig}^N| = \frac{d^{N+1}-1}{d-1}$}. 

We set-up a signature transform for predicting the future observations in a healthcare setting as shown in Figure~\ref{fig:dst_arch}. 
We pass the sequence of observations $\tau_{j,0:t} = \{\Ob_0,\dots,\Ob_t\}$ up to the current time through a pointwise encoder $\psi$. 
The resulting sequence $\psi(\tau_{j,0:t})$ is processed by the signature transform $\text{Sig}^N(\psi(\tau_{j,0:t}))$ of order $N$. This sequence is then passed through a recurrent neural network to produce the learned state representation $\hat\St_{0:t}$. This state representation is then passed through the decoder $\phi$ to predict the SO $\hat\Ob_{t+1}$.

Implemented using the Signatory library\footnote{\url{https://github.com/patrick-kidger/signatory} }, this is essentially equivalent to using the signature transformation as a stream preserving non-linear transformation layer in a neural network.

Recently, signature transforms have been incorporated into modern neural network architectures and have been shown to have great promise in a variety of learning paradigms~\citep{bonnier2019deep}.
Notably, a model architecture utilizing a signature transform for sepsis prediction won the 2019 Physionet challenge~\citep{morrill2019signature}.
The success of such a model demonstrates that such transforms may be capable of adequately handling irregularly sampled time-series data from partially observable environments.

In the encoder, we start with two pointwise one-dimensional convolutional layers (with a kernel size of 1) to add 8 augmented features to the 63 dimensional input vector. We then apply a stream preserving signature transformation with a depth of 2. The latent states are obtained by passing the output of the signature transform through a 2 layer GRU with $dim$ hidden units, where $dim$ is the chosen embedding dimension.

For estimating the decoder function $\phi$, we again use two pointwise one-dimensional convolutional layers (with a kernel size of 1) with filter sizes of 64 and 32 respectively. Then, we apply a stream preserving signature transformation of depth 2. Finally, We use a pointwise 2-layer feed-forward neural network with sizes $(|Sig^N|, 64)$, and $(64,33)$ with ReLU activation.

The best DST architectures for each choice of $\hat\dimSt$ were trained for 50 epochs with a learning rate of $10^{-3}$. The $\lambda$s for regularizing the training to correlate with acuity scores are set to 1.

\subsection{ODE-RNN}
\label{sec:apdx_archODE}

\begin{figure}
\centering
\includegraphics[width=\linewidth]{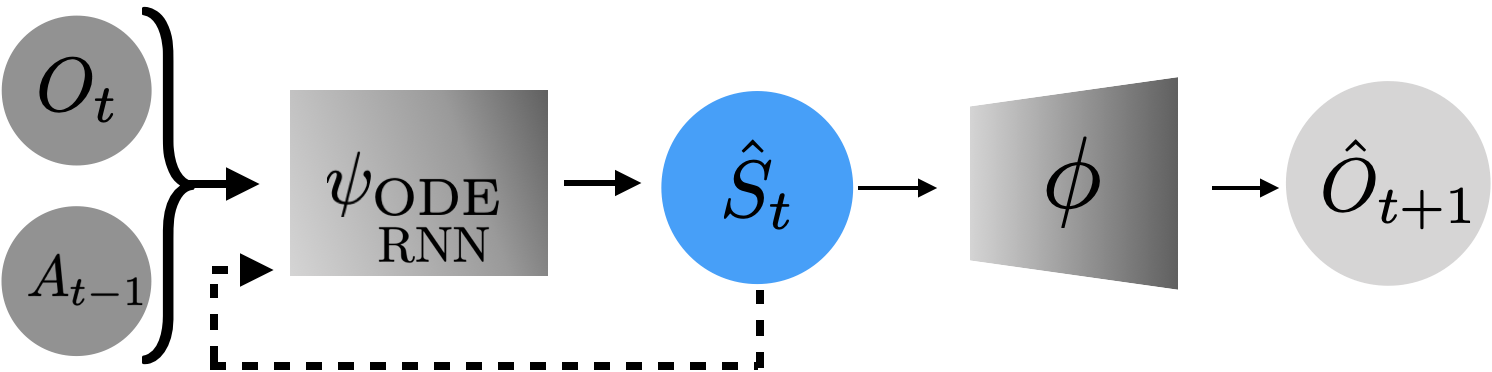}
\caption{\footnotesize The ODE-RNN architecture for SO prediction}
\label{fig:ode_arch}
\end{figure}

\cite{rubanova2019latent} generalize the latent transitions between observations inside an RNN to a continuous time differential equation using neural networks, building from the Neural ODE~\citep{chen2018neural} framework. An ODE-RNN is a recurrent neural network where the hidden states between observations evolve according to a parameterized ODE\footnote{To implement the ODE-RNN, we use the code available at \url{https://github.com/YuliaRubanova/latent_ode}}.



Although ODE-RNNs are natively able to handle missing values and irregularly sampled time series, for the purposes of this paper, we still use imputed, time-binned data for this model.

For the encoder, we use a GRU with 50 units, where the hidden states between observations are modelled by a Neural ODE parameterized by a 2-layer MLP with 50 hidden units. We use the adaptive stepsize \texttt{dopri5} solver. For the decoder, we use an MLP applied at each time step, consisting of 3 layers, with sizes $(dim, 100)$, $(100, 100)$, $(100, 33)$ and ReLU activations.

The best ODE-RNN architectures for each setting of $\hat\dimSt$ were trained for 100 epochs with a learning rate of $10^{-3}$. The $\lambda$s for regularizing the training to correlate with acuity scores are all set to 1.

Note that though latent ODE representations have shown better performance in representing time series data, we do not believe that a latent ODE is appropriate for this task. This is because the encoder of a latent ODE involves an ODE-RNN running backwards in time over the inputs to obtain a probability distribution over $z_0$. Thus, the initial latent state would contain information about all subsequent observations and actions (similar to if a bidirectional RNN were used). A sampled value of $z_0$ is then used as the initial condition in a Neural ODE to solve for $z_{1:T}$.  This information leakage would result in an unrealistic estimate of the SO prediction error.

\subsection{CDE}
\label{sec:apdx_archCDE}

\begin{figure}
\centering
\includegraphics[width=\linewidth]{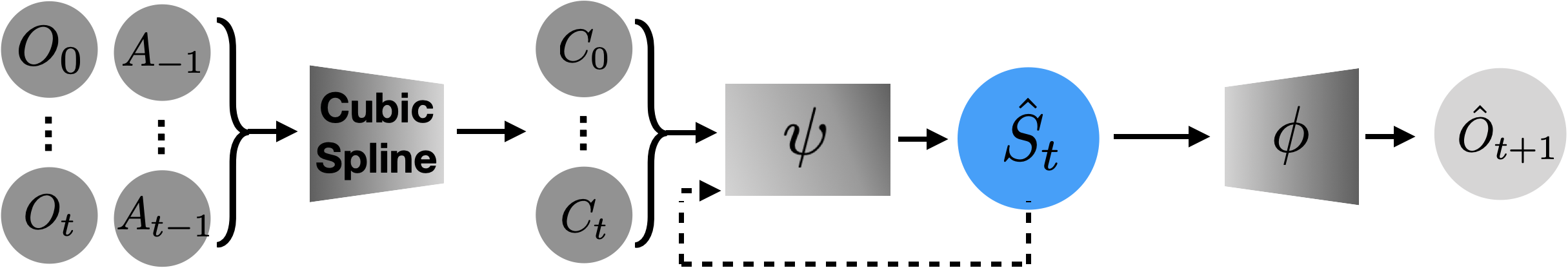}
\caption{\footnotesize The Neural CDE architecture for SO prediction}
\label{fig:cde_arch}
\end{figure}

Similar to ODE-RNNs \citep{rubanova2019latent}, Neural Control Differential Equations (CDEs) \citep{kidger2020neural} model temporal dynamics by parameterizing the time derivative of the hidden states by a neural network. Unlike ODE-RNNs, the hidden states in CDEs evolve smoothly as a function of time, even at time points when data is observed. To accomplish continual dependence on the data throughout the latent trajectory, cubic spline interpolation is used, and the network operates on pre-computed cubic spline coefficients instead of the actual observations. The initial value for the latent space is calculated by a linear map on the inputs at $t=0$. It has been shown that CDEs are universal approximators from sequences in $\mathbb{R}^D$ to real valued targets. 

For the encoder, we use a Neural CDE parameterized by an MLP with four hidden layers, each of which has 100 hidden units. We use ReLU activation for the hidden layers, and a tanh activation for the final layer. For the decoder, we use an MLP applied at each time step, consisting of 3 layers, with sizes $(dim, 100)$, $(100, 100)$, $(100, 33)$ and ReLU activations.

The best CDE architectures for each setting of $\hat\dimSt$ were trained for 200 epochs with a learning rate of $2\times10^{-4}$. The $\lambda$s for regularizing the training to correlate with acuity scores are all set to 1.

\subsection{Autoencoder}
\label{sec:apdx_archAE}

\begin{figure}
\centering
\includegraphics[width=0.85\linewidth]{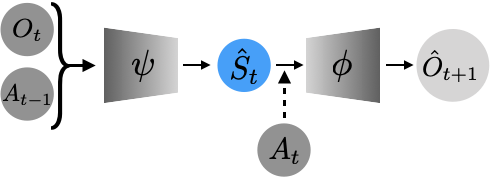}  
\caption{\footnotesize The Autoencoder architecture}
\label{fig:ae_arch}
\end{figure}

To isolate the contribution of the recurrent layer in the RNN (Sec.~\ref{sec:apdx_archRNN}), we also evaluate a simple autoencoder that replaces that layer in the encoding function $\psi$ with a fully connected layer to produce the state representation $\hat{\St_t}$. As is done with AIS (Sec.~\ref{sec:apdx_archAIS}), we concatenate the current action $\Ac_t$ to $\hat{\St}_t$ when predicting the SO $\hat{\Ob}_{t+1}$ using the decoder function $\phi$. The autoencoder architecture shown in Figure~\ref{fig:ae_arch} was trained using same loss function as the RNN, AIS, and DST approaches.

The autoencoder's encoding function $\psi$ is comprised of a three layer fully connected neural network with ReLU activations with sizes $(58,64)$, $(64, 128)$, $(128,\hat\dimSt)$ to produce the state representation $\hat\St_t$. To produce an approximation of the SO, $\hat\St_t$ is concatenated with the current action $\Ac_t$ and passed to the decoding function $\phi$, another three layer fully connected neural network with ReLU activations. The sizes of the layers comprising $\phi$ are $(\hat\dimSt+25,64)$, $(64,128)$ and $(128,33)$. We train this model end-to-end using the loss function, with the option to be regularized by the correlation coefficient.

The best autoencoder models for each setting of $\hat\dimSt$ were trained for 600 epochs with a learning rate of $5e-4$. The $\lambda$s for regularizing the training were all set to 100.

\section{State construction in prior work}
\label{sec:apdx_prior_table}

See Table~\ref{tab:background} for an overview of how prior work has constructed state representations for RL in healthcare settings.

\begin{table*}[ht]
\caption{\small State construction for RL in healthcare - background}
\label{tab:background}
\begin{tabular}[t]{m{0.25\linewidth}m{0.35\linewidth}m{0.4\linewidth}}
\toprule
\small
 \textbf{Ref} & \textbf{Domain} & \textbf{State Construction} \\ \midrule
  \cite{hauskrecht2000planning} & Heart Disease Management &  10 categorical variables + alive/dead; constructed hierarchically \\ \midrule
  \cite{guez2008adaptive} & Epilepsy  & 114 dimensional continuous - summarizing past EEG activity\\ \midrule
  \cite{shortreed2011informing} & Schizophrenia Treatment & 20 demographic + 30 time varying; imputed using  fully conditional specification  \\ \midrule
  \cite{tsoukalas2015data} & Sepsis & 9 states constructed from vitals based on medical criteria \\ \midrule
  \cite{nemati2016optimal} & Medication dosing & Estimated using discriminative hidden Markov model on 21 continuous vitals + 6 binary demographics \\ \midrule
  \cite{raghu2017deep} &  Sepsis (MIMIC-III) & Time augmented last observation (47 + 1 = 48 dimensional)  \\ \midrule
  \cite{prasad2017reinforcement}  &  Weaning of mechanical ventilation (MIMIC-III) & Last observation (32 dimensional)  \\ \midrule
  \cite{parbhoo2017combining} & HIV Treatment & 7 hidden discrete physiological  states from Bayesian model-based RL over 80 observations \\ \midrule
  \cite{KomorowskiCeliBadawiEtAl2018} & Sepsis (MIMIC-III)  & Clustered state with 750 clusters \\ \midrule
  \cite{raghu2018model} & Sepsis (MIMIC-III)  & $k$-Markov with $k=4$; 
  $198 = 4 \times 47$ dimensional state space\\ \midrule
  \cite{li2018actor} & Sepsis (MIMIC-III) & 5 demographic + 46 time varying ; modelled as Gaussian mixture \\ \midrule
  \cite{peng2018improving} & Sepsis (MIMIC-III) & Sequence embedding with RNN (128 dimensional hidden state from 43 features) \\ \midrule
  \cite{chang2019dynamic} & Sepsis (MIMIC-III) & Last observation (39 dimensional extracted from time-series + 38 static covariates)  \\ \midrule
  \cite{cheng2019optimal} & Lab testing (MIMIC-III) & Last observation (21 dimensional). Data imputation done using a Multi-output Gaussian Process framework.\\ \midrule
  \cite{li2019optimizing} & Sepsis (MIMIC-III) & Auto-encoding SMC over 48 patient variables \\\midrule
\bottomrule
\end{tabular}
\end{table*}

\section{Additional experimental results}
\label{sec:apdx_results}
In this section we include a more exhaustive accounting of the experimental results that did not fit within the space constraints of the main body of the paper.

\subsection{Policy Training}
\label{sec:apdx_policy_training}

\begin{figure*}
\centering
\includegraphics[width=\textwidth]{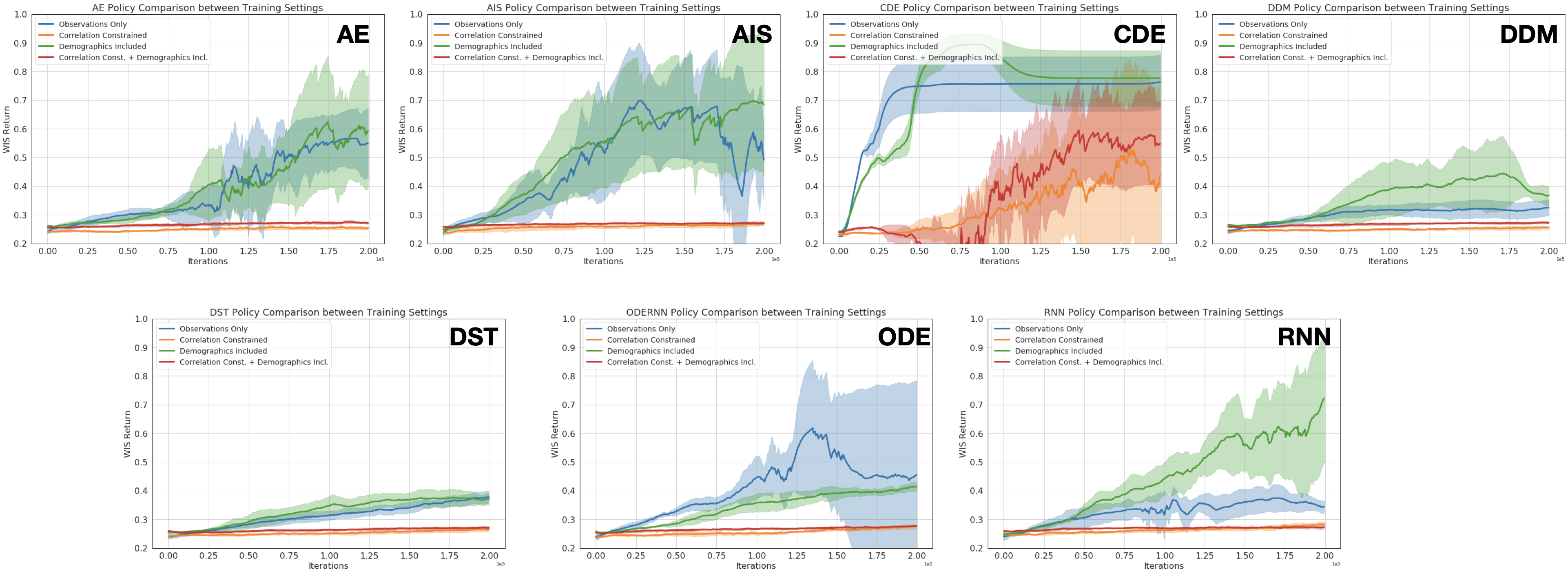}
\caption{\footnotesize A compilation of the policy learning curves for each representation training setting for all encoding approaches investigated in this paper.}
\label{fig:apdx_policy_training}
\end{figure*}

We train policies on each of the learned state representations outlined in Section~\ref{sec:models}. As we do not have the ability to generate more data through an exploration of novel treatment strategies, we develop a policy using offline, batch reinforcement learning. In this setting, it is critical that the estimated value function not extrapolate to actions that are not present in the provided data~\citep{gottesman2019guidelines}. To counter this error caused by extrapolation, \citet{fujimoto2019off} developed an algorithm for continuous control settings that truncates any Q-function estimate corresponding to actions that fall outside the support of the dataset. This algorithm, Batch Constrained Q-Learning (BCQ) was then adapted and simplified by the authors for use in discrete action settings~\citep{fujimoto2019benchmarking}.

As the patient cohort that we have to learn policies from is defined with discrete actions, we use the simplified Batch Constrained Q-Learning (BCQ) for discrete action settings~\citep{fujimoto2019benchmarking} to learn treatment policies from state representations $\hat\St$. We train the policies using the encoded training subset of our data, validating the performance of the policy using the testing subset via weighted importance sampling (WIS), following~\cite{li2019optimizing}. The WIS return for each policy throughout training is computed by: $R^{\mathrm{WIS}} = \frac{\sum_n^N w_n R_n}{\sum_n^N w_n}$, where the $w_n$ are the per-trajectory $\mathrm{IS}$ weights and $R_n$ is the observed outcome of the trajectory.

WIS evaluation of policies trained from the representations encoded by the architectures outlined in Section~\ref{sec:methods}. The Q-network used in our implementation of BCQ was comprised of 3 fully connected layers, using 64 nodes per layer (excepting for the DDM architecture where we used 128 nodes per layer). The learning rate was empirically tuned for each training approach in a log-uniform range of $\{1e-5, 1e-2\}$. The best policies for all approaches, excepting CDE, used a learning rate of $1e-3$. CDE used a learning rate of $1e-5$. The BCQ action eliminiation threshold $\tau$ was set to $0.3$ for all experiments.

All policies were trained\footnote{We adpated~\cite{fujimoto2019benchmarking}'s code which can be accessed at: \url{https://github.com/sfujim/BCQ/tree/master/discrete_BCQ}} from a uniformly sampled experience replay buffer comprised of the training batch of patient trajectories for 200k iterations, evaluating the trained policy every 500 iterations using the testing subset of the patient data. 

The behavior policy used in WIS was derived via behavior cloning using a two layer fully connected neural network trained with a supervised cross entropy loss using the stored actions with corresponding actions. WIS evaluation was performed by using the observation drawn from the test set of the patient data, predicting the observed action using the approximated behavior policy and then comparing with the inferred action provided by the current policy trained with BCQ using the corresponding state representation encoded by the user's choice of information encoding approach. 

In Figure~\ref{fig:apdx_policy_training}, we present the evaluations of policies learned from the representations learned through each information encoding approach. Each subfigure features the policy performance based on the training strategy used to learn the state representations.

\subsection{Analysis of correlation coefficient between representations and acuity scores}
\label{sec:apdx_correlation}

\begin{figure*}
\centering
\includegraphics[width=\textwidth]{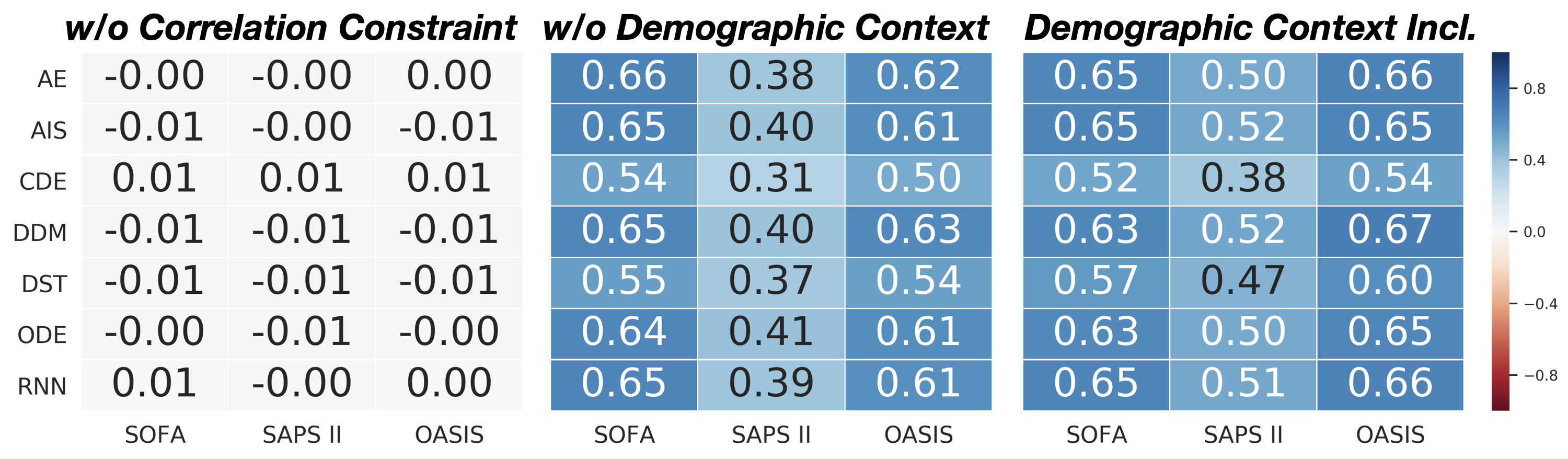}
\caption{The average Pearson correlation coefficient between the state representations from each encoding approach and acuity scores. Shown here are the average coefficients when the representation learning process is unregularized (left), when demographic features are omitted (center) or are included (right). The inclusion of demographic features when constructing the state representations causes them to be more correlated. When the state representations are uncoorelated. They fail to embed information directly correlated with the derived acuity scores.}
\label{fig:apdx_corrHtmp}
\end{figure*}

Here we present in Figure~\ref{fig:apdx_corrHtmp} the average correlation coefficients between the acuity scores and learned state representations from the various information encoding approaches. What is compared here is the effect of representation learning process on the subsequent correlation coefficients.

\subsection{PCA Figures}
\label{sec:apdx_pca}

This section contains the nonlinear projection using PCA of the state representations learned from each approach. For simplicity, we only include the representations for the first and final observations of each patient trajectory, colored by the corresponding SOFA score. We also draw lines connecting these points to help infer how the patient's health evolves, as demonstrated in representation space. To aid this inference, we've colored the lines according to patient outcome. Blue lines signify patients who overcame sepsis and survived. Red lines connect the observations of those patients who died following complications associated with their sepsis diagnosis.

\begin{figure*}
    \centering
    \includegraphics[width=\textwidth]{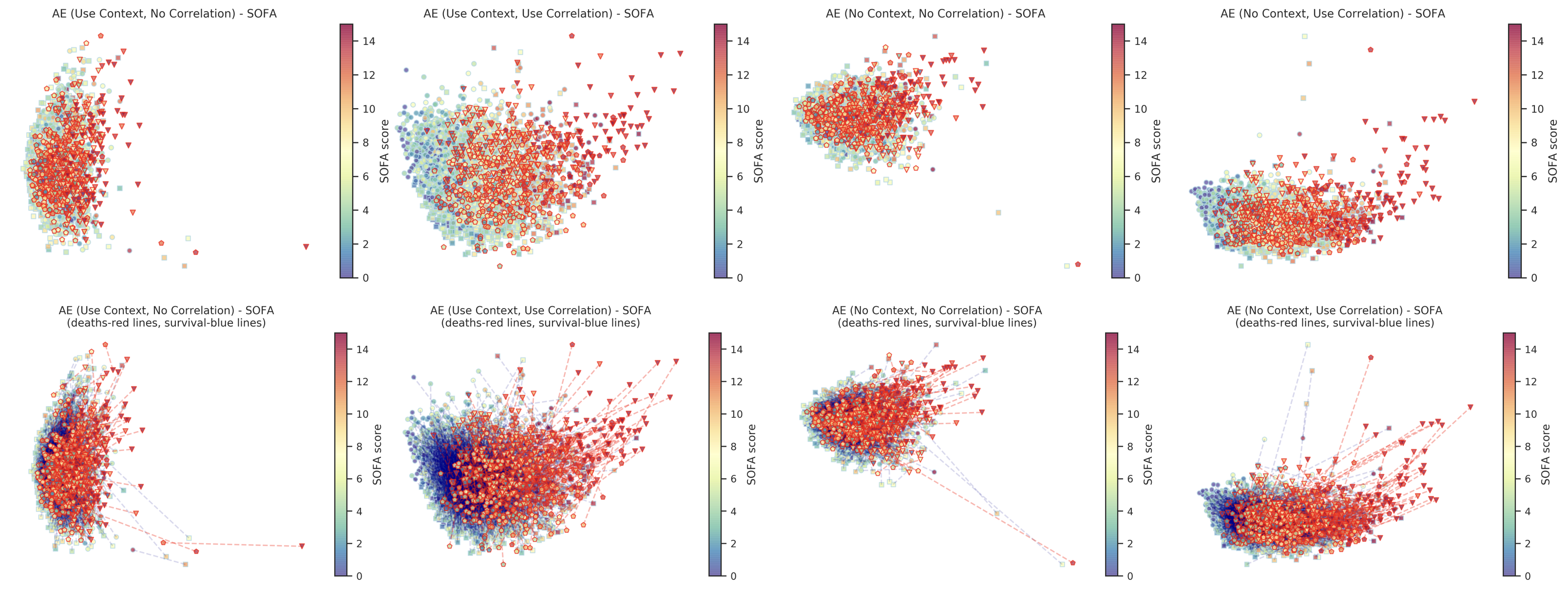}
    \caption{\footnotesize Representations of patient health, learned through an Autoencoder (AE)}
    \label{fig:apdx_ae_pca}
\end{figure*}

\begin{figure*}
    \centering
    \includegraphics[width=\textwidth]{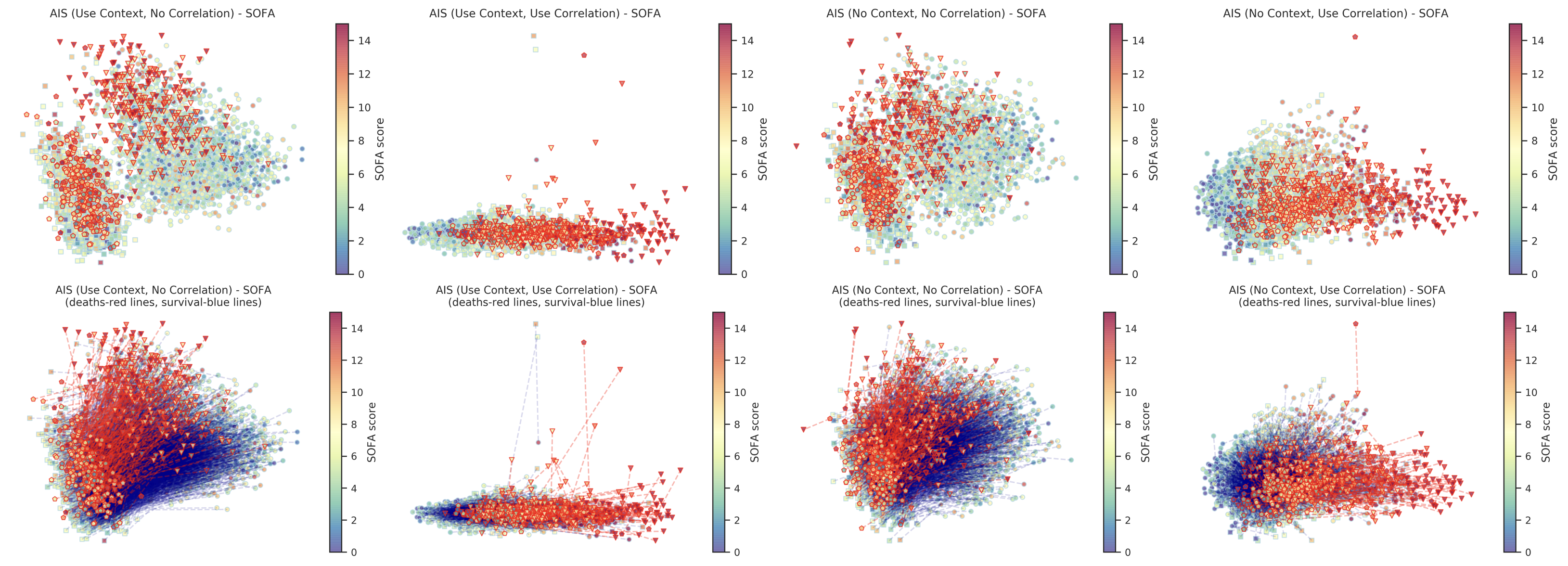}
    \caption{\footnotesize Representations of patient health, learned through Approximate Information State (AIS)}
    \label{fig:apdx_ais_pca}
\end{figure*}

\begin{figure*}
    \centering
    \includegraphics[width=\textwidth]{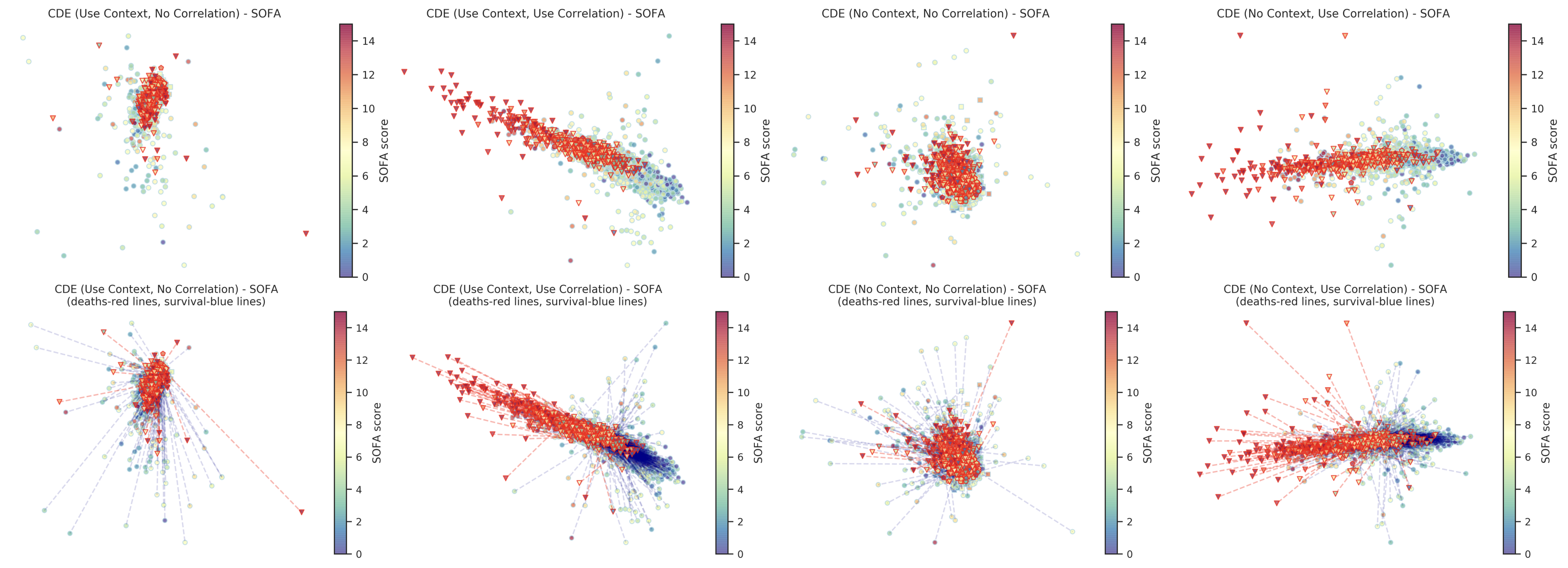}
    \caption{\footnotesize Representations of patient health, learned through the Neural CDE (CDE)}
    \label{fig:apdx_cde_pca}
\end{figure*}

\begin{figure*}
    \centering
    \includegraphics[width=\textwidth]{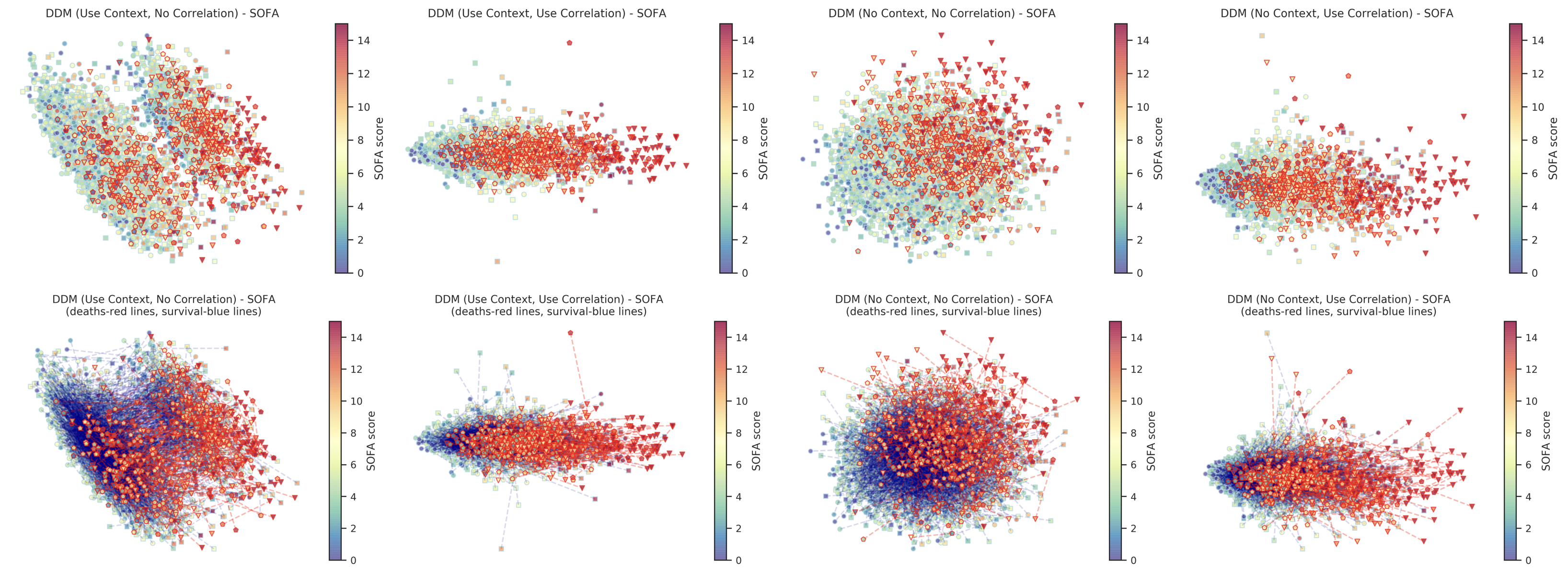}
    \caption{\footnotesize Representations of patient health, learned through the Decoupled Dynamics Module (DDM)}
    \label{fig:apdx_ddm_pca}
\end{figure*}

\begin{figure*}
    \centering
    \includegraphics[width=\textwidth]{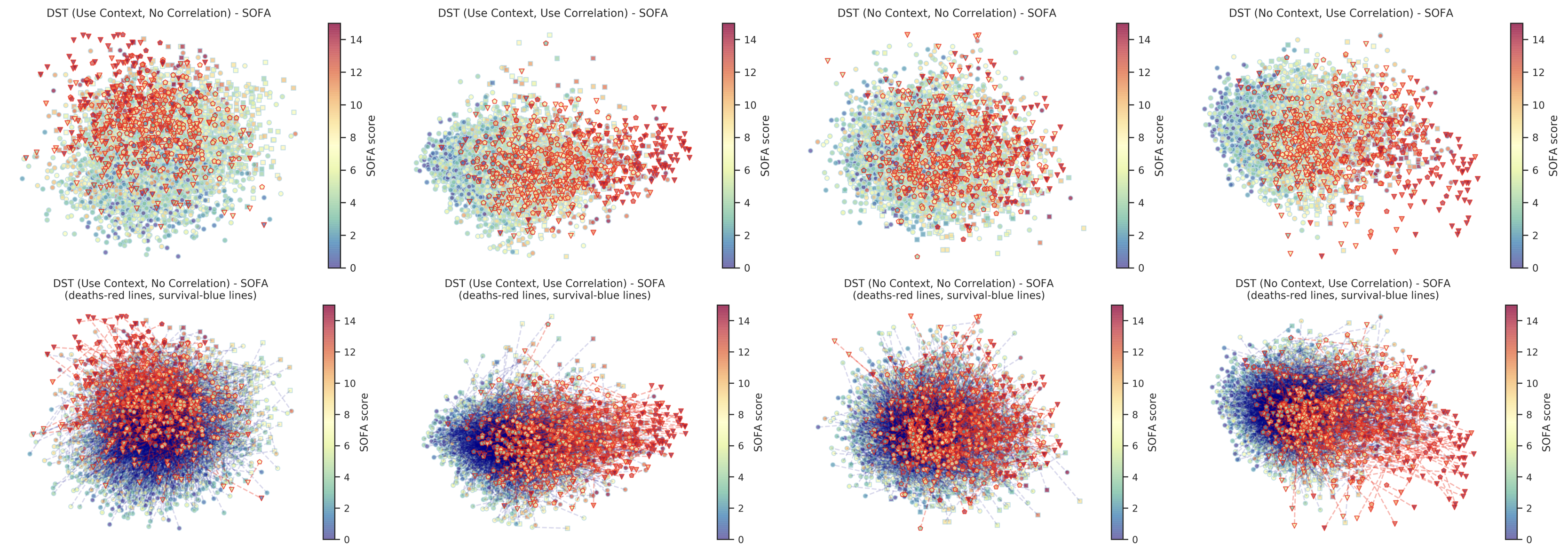}
    \caption{\footnotesize Representations of patient health, learned through the Deep Signature Transform (DST)}
    \label{fig:apdx_dst_pca}
\end{figure*}

\begin{figure*}
    \centering
    \includegraphics[width=\textwidth]{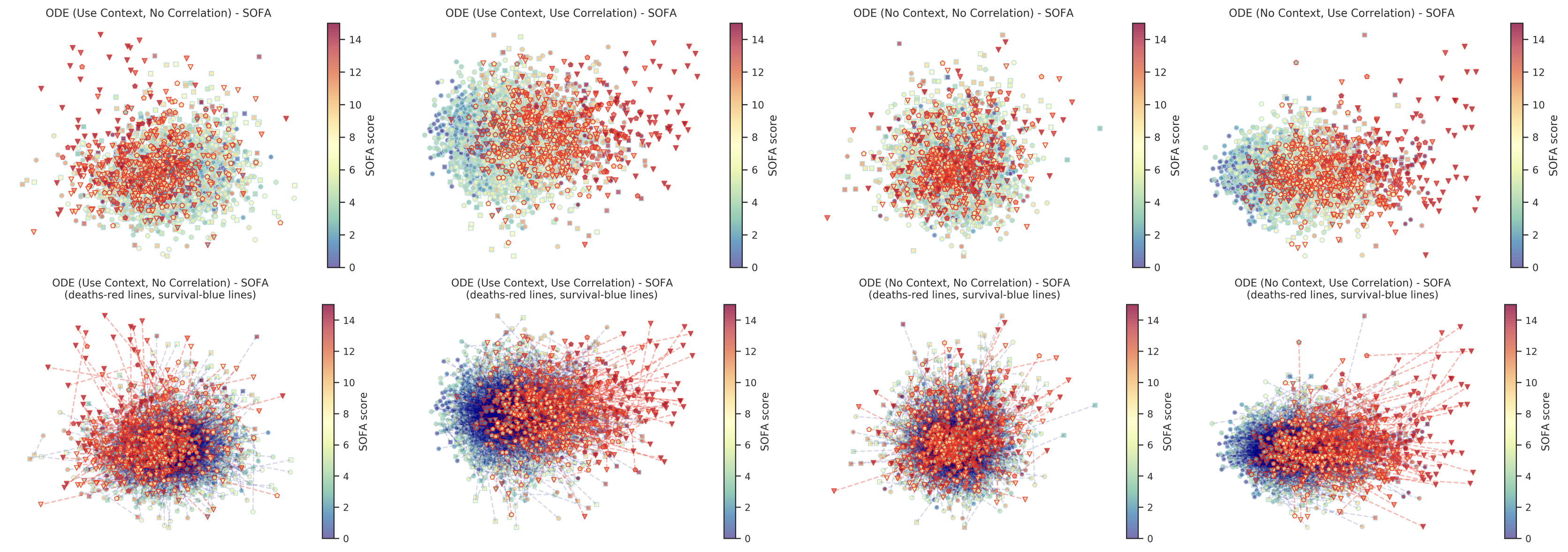}
    \caption{\footnotesize Representations of patient health, learned through the ODE-RNN (ODE).}
    \label{fig:apdx_ode_pca}
\end{figure*}

\begin{figure*}
    \centering
    \includegraphics[width=\textwidth]{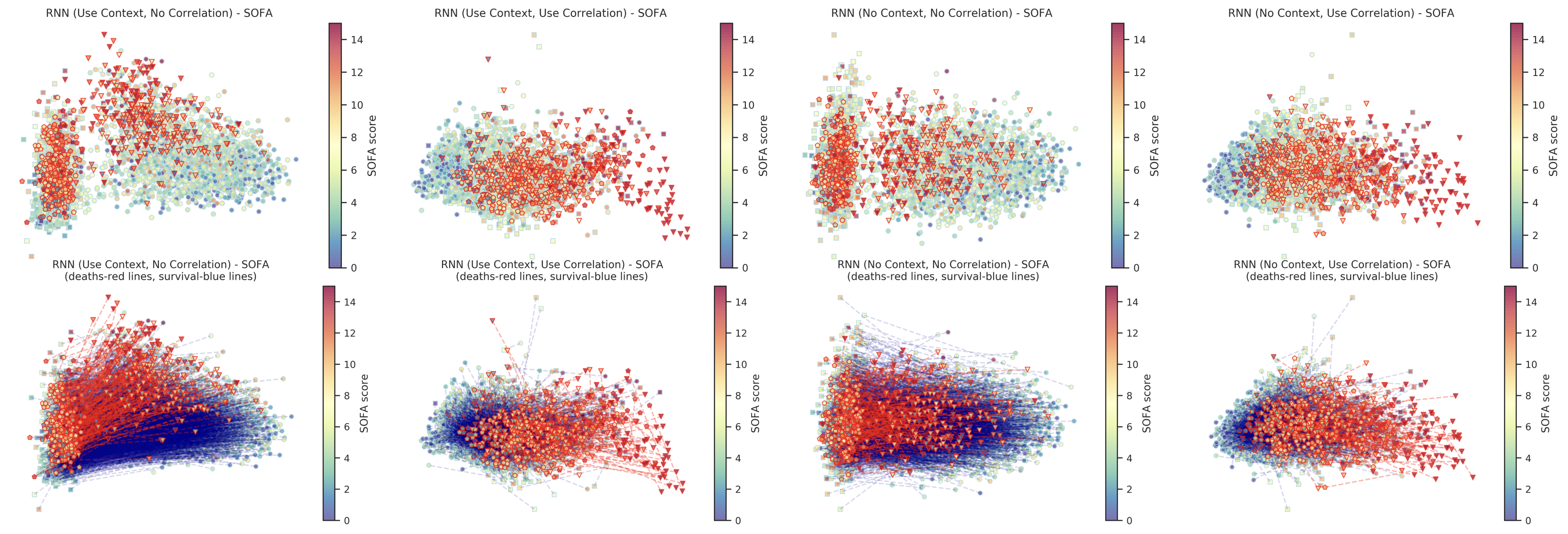}
    \caption{\footnotesize Representations of patient health, learned through a recurrent autoencoder (RNN)}
    \label{fig:apdx_rnn_pca}
\end{figure*}

\end{document}